\documentclass[letterpaper]{article} 
\usepackage{aaai2026}  
\usepackage{times}  
\usepackage{helvet}  
\usepackage{courier}  
\usepackage[hyphens]{url}  
\usepackage{graphicx} 
\usepackage{natbib}  
\usepackage{caption} 
\usepackage{algorithm}
\usepackage{algpseudocode}
\usepackage{amsfonts}
\usepackage{booktabs}
\usepackage{multirow}
\usepackage[table]{xcolor} 
\usepackage{subcaption}
\usepackage{xcolor}      
\usepackage{amsmath,amssymb}    

\usepackage{newfloat}
\usepackage{listings}
\DeclareCaptionStyle{ruled}{labelfont=normalfont,labelsep=colon,strut=off} 
\floatstyle{ruled}
\newfloat{listing}{tb}{lst}{}
\floatname{listing}{Listing}
\usepackage{newunicodechar}
\newunicodechar{✓}{\checkmark} 

\definecolor{mypink}{rgb}{1.0,0.8,0.8}
\definecolor{myblue}{rgb}{0.8,0.8,1.0}
\definecolor{mygreen}{rgb}{0.8,1.0,0.8}
\definecolor{myorange}{rgb}{1.0,0.9,0.8}

\title{Mask the Redundancy: Evolving Masking Representation Learning for Multivariate Time-Series Clustering}

\author {
    Zexi Tan,
    Xiaopeng Luo,
    Yunlin Liu,
    Yiqun Zhang\thanks{Corresponding Author}
}
\affiliations{
    School of Computer Science and Technology, Guangdong University of Technology, Guangzhou, China\\
    3123004194@mail2.gdut.edu.cn, luoxiaopeng@mails.gdut.edu.cn, lyunlin2713@gmail.com, yqzhang@gdut.edu.cn
}

\begin{document}

\maketitle

\begin{abstract}
Multivariate Time-Series (MTS) clustering discovers intrinsic grouping patterns of temporal data samples. Although time-series provide rich discriminative information, they also contain substantial redundancy, such as steady-state machine operation records and zero-output periods of solar power generation. Such redundancy diminishes the attention given to discriminative timestamps in representation learning, thus leading to performance bottlenecks in MTS clustering. Masking has been widely adopted to enhance the MTS representation, where temporal reconstruction tasks are designed to capture critical information from MTS. However, most existing masking strategies appear to be standalone preprocessing steps, isolated from the learning process, which hinders dynamic adaptation to the importance of clustering-critical timestamps. Accordingly, this paper proposes the Evolving-masked MTS Clustering (EMTC) method, whose model architecture comprises Importance-aware Variate-wise Masking (IVM) and Multi-Endogenous Views (MEV) generation modules. IVM adaptively guides the model in learning more discriminative representations for clustering, while the reconstruction and cluster-guided contrastive learning pathways enhance and connect the representation learning to clustering tasks. Extensive experiments on 15 benchmark datasets demonstrate the superiority of EMTC over eight SOTA methods, where the EMTC achieves an average improvement of 4.85\% in F1-Score over the strongest baselines.
\end{abstract}

\begin{links}
    \link{Code}{https://github.com/yueliangy/EMTC}
\end{links}

\section{Introduction}
Multivariate Time-Series (MTS) clustering~\cite{PRMTSC21,Zhang23MTSC} is a pivotal unsupervised data analysis task~\cite{zhang2025learning,tan2025meetsepsismultiendogenousviewenhancedtimeseries} for discovering intrinsic patterns of temporal data, which are common in domains like activity recognition~\cite{ACM21MaMTSC}, industrial monitoring~\cite{ALWAN2022MTSC}, and medical data analysis~\cite{xie2025de3sdualenhancedsoftsparseshapelearning}. Although MTS contains rich trend and periodic information~\cite{Zhang2023Learning}, widespread redundant timestamps may obscure the representation of key distribution patterns, thereby compromising the formation of compact and meaningful cluster structures. Conventional approaches implicitly mitigate MTS redundancy through hand-crafted feature engineering, dimensionality reduction~\cite{YangMMDB04,LI2019239}, and elastic time-series alignment~\cite{LI2020105907,CaiAccess25}. Despite their effectiveness in certain scenarios, they often impose strong temporal shape or sample distribution assumptions, and fail to support end-to-end representation learning.

\begin{figure}[!t]
\centering
\includegraphics[width=\columnwidth]{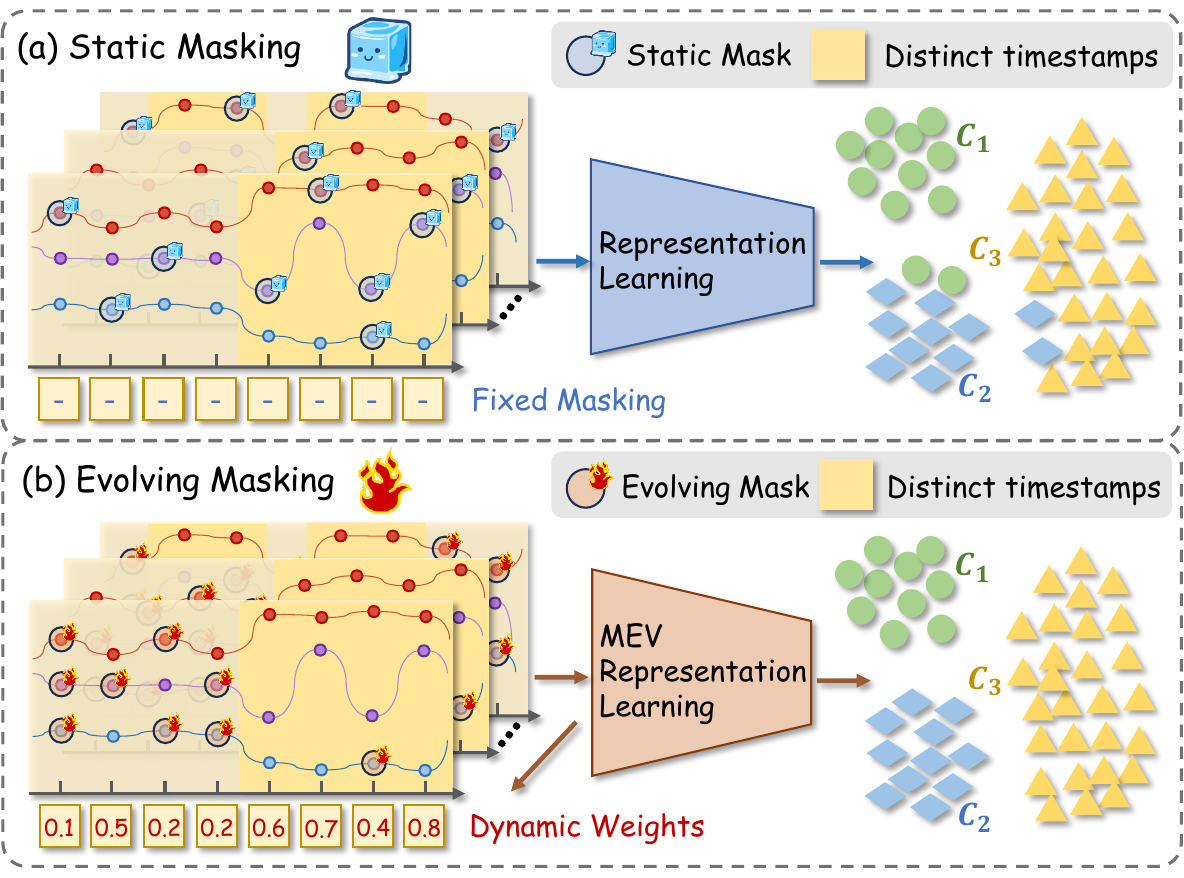}
\caption{\textbf{(a) Static Masking vs. (b) Evolving Masking (ours).} Static masking is a fixed MTS preprocessing strategy to enhance the downstream representation learning. In contrast, our evolving masking dynamically adapts to the representation learning and clustering objective through attention-based timestamp weighting, simultaneously surpassing the MTS redundancy and enhancing the cluster discrimination of representations.}
\label{motivation}
\end{figure}

Deep learning paradigms~\cite{Lafabregue2021Endtoend,IENCO202336} for MTS clustering usually learn representations and then feed the outputs to algorithms like $k$-means~\cite{MacQueen1967SomeMF,IKOTUN2023178} to obtain the final clustering results. In this stream, the Autoencoder-based frameworks~\cite{Ienco2020Deep,li2023timaeselfsupervisedmaskedtime} are common, which derive latent high-quality embeddings through reconstruction-guided model training. However, their emphasis on reconstruction fidelity may act to retain the MTS redundancy. Moreover, since the clustering objective is usually separated from the representation learning, the learning process fails to suppress the redundancy from the clustering-friendly perspective. Accordingly, contrastive learning methods~\cite{Wu24MM,LI2025102812,wang2025fcacc1} have been employed to learn discriminative features by contrasting positive and negative instance pairs. Although it has been proven that the self-supervised contrastive learning can implicitly obtain clustering-friendly representations~\cite{Yann2023NeurIPS}, their effectiveness critically depends on the construction of instance relationships. That is, when the contrastive strategies do not align with the sought cluster distributions, the learned representations may preserve or even amplify temporal redundancies.

To explicitly address the temporal redundancy, attention~\cite{Ienco2020Deep} has been adopted to perform soft redundancy filtering through dynamic timestamp weighting~\cite{DING2023527}. Although this type of method considerably relieves the impact of redundancy and enhances the representation of timestamps, it also inherently preserves the full input structure containing the redundant information. More importantly, the learned attention weights may be misled by highly activated yet non-informative patterns, thus failing in prioritizing sparse-but-critical features. In contrast, masking mechanisms offer a more structured approach to redundancy suppression in MTS representation learning~\cite{TACTiS,EmadeldeenICML24,Zhang24reconmask}. Conventional static masking~\cite{FuIJCNN22,zhang2023time} compels models to learn from partial observations through reconstruction. Although masking has been well validated in acquiring general representations, the fixed schemes are incompetent in task adaptation as demonstrated in Figure~\ref{motivation} (a). Most recently, dynamic masking~\cite{Li23CVPR,shi2025trainabledynamicmasksparse} has been developed to learn the masking. Nevertheless, learning masks for MTS redundancy suppression and structural discriminative representation remains underexplored.

This paper, therefore, proposes Evolving-masked MTS Clustering (EMTC) to explicitly address the critical redundancy issue in MTS clustering. As illustrated in Figure~\ref{motivation} (b), EMTC adopts attention-based Importance-aware Variate-wise Masking (IVM) to dynamically determine the redundant timestamps to be excluded in the current learning epoch, and introduces Multi-Endogenous Views (MEV)-based representation learning to facilitate a more informative and generalized learning process. More specifically, the IVM module assigns masks to timestamps with low importance to explicitly prevent them from participating in the representation learning, where the attention-based importance is learned for each time series. Then, the MEV is generated based on the masked MTS, and a dual-path architecture is designed to learn representations based on the MEV: 1) Consistency and Reconstruction Learning (CRL) performs intra- and inter-view reconstruction to extract invariant features that are robust to masked redundancy and view-specific information, respectively, and 2) Clustering-guided MEV Contrastive learning (CMC) adopts clusters as a basis for data augmentation, which serves to enhance the cluster separation in the embedding space and also connects the clustering objective into representation learning. The contributions are as follows:
\begin{itemize}
    \item \textbf{A novel learnable redundancy masking mechanism.} This paper designs a content-aware attention mechanism for timestamp scoring, which dynamically guides the evolving masking of redundant timestamps. This directly and adaptively suppresses temporal redundancy, enhancing representation discriminability for MTS clustering.
    \item \textbf{Synergistic design of IVM-MEV complementary.} A complementary redundancy masking and multi-view learning mechanism has been established, where MEV facilitates sufficient interaction of the MTS to avoid the dominance of the crisp IVM masking, while IVM serves to eliminate the redundancy amplified by the MEV.
    \item \textbf{Representation and clustering connected paradigm.} EMTC is an early exploration that integrates contrastive learning into MTS clustering. By leveraging dynamically updated clustering results to guide data augmentation, it connects both the representation learning and the redundancy masking learning to the MTS clustering objective.
\end{itemize}

\section{Related Work}

\begin{figure*}[ht]
\centering
\includegraphics[width=\textwidth]{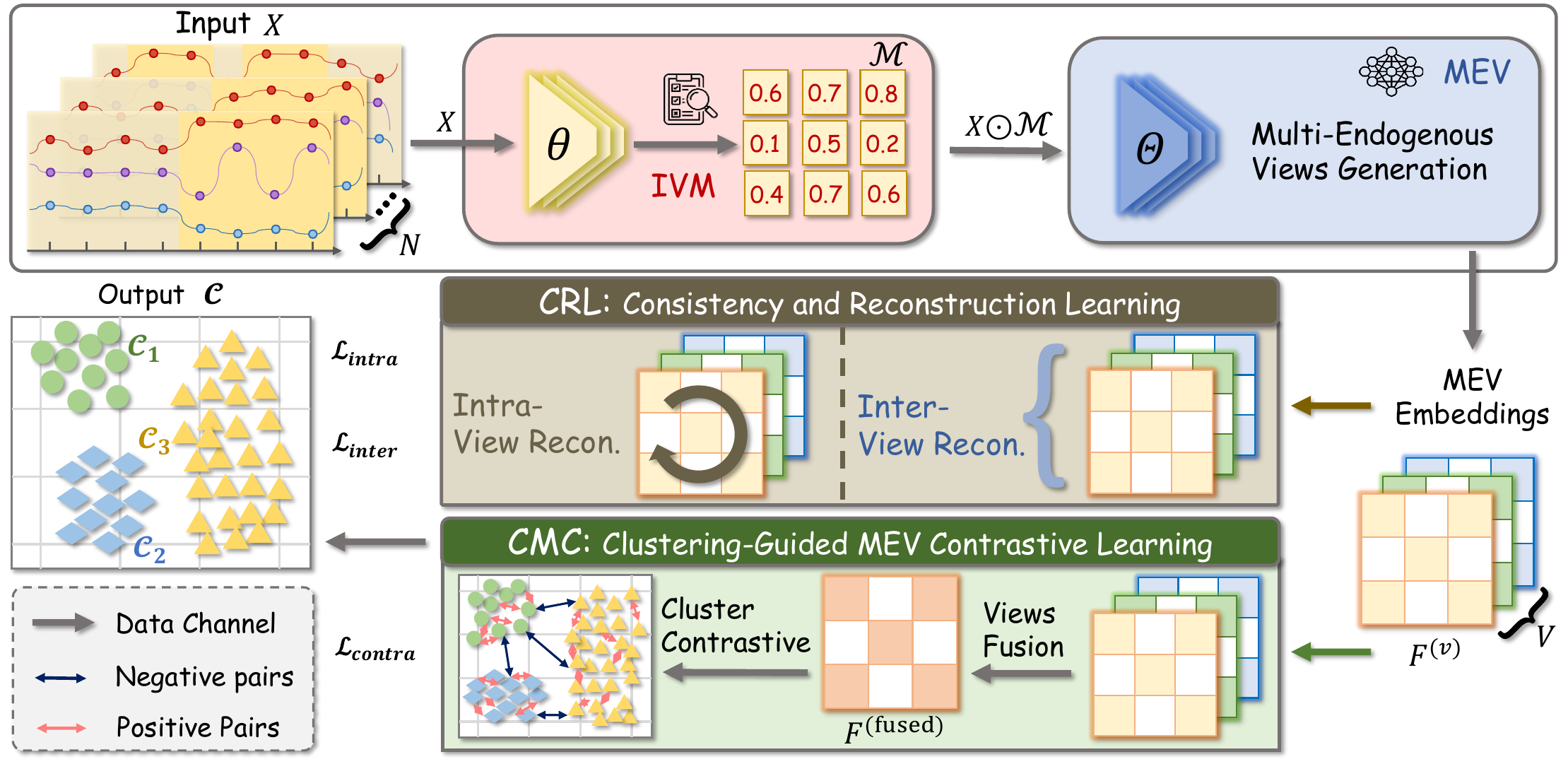}
\caption{Overview of the proposed EMTC framework.} \label{Overview}
\end{figure*}

\textbf{Redundancy Suppression in MTS Clustering.} Traditional MTS clustering tackles redundancy via feature engineering or linear dimensionality reduction (e.g., PCA variants)~\cite{YangMMDB04,BARRAGAN2016144,LI2019239,SihoKDD22}, projecting data into lower-dimensional spaces but constrained by linear assumptions. Dynamic Time Warping (DTW) methods~\cite{LI2020105907,SHEN2023611,CaiAccess25} offer adaptive alignment but lack dynamic redundancy suppression. Sample partition algorithms like $k$-means~\cite{MacQueen1967SomeMF,Ukmeans,IKOTUN2023178} and hierarchical clustering~\cite{NIPS2003_23483314,ICDM2007,mhcclaaai} rely on predefined metrics and are sensitive to initialization. Deep learning paradigms learn discriminative representations end-to-end. Autoencoder frameworks like DeTSEC~\cite{Ienco2020Deep} and RDDC~\cite{ICASSP19Daniel} derive compact embeddings, while contrastive methods (TimesURL~\cite{Liu_Chen_2024}, FCACC~\cite{wang2025fcacc1}, MVCIMTS~\cite{LI2025102812}) enhance feature distinctiveness through augmented views, though efficacy depends on augmentation design. Other innovations include CDCC~\cite{Peng_Luo_Lu_Wang_Li_2024} for cross-domain contrastive learning, TimeDRL~\cite{ChangICDE} for disentangled representation learning, and $k$-Graph~\cite{TKDE2025Boniol} for interpretability. Although competent in capturing complex temporal structures, most lack explicit dynamic mechanisms to suppress evolving temporal redundancy.

\textbf{Masking Strategies for MTS Analysis.} The principle of masked learning demonstrates significant benefits across domains like computer vision and NLP, enhancing efficiency and precision. For instance, DynaMask~\cite{Li23CVPR} dynamically selects optimal mask resolution in instance segmentation, while Trainable Dynamic Mask Sparse Attention~\cite{shi2025trainabledynamicmasksparse} achieves adaptive sparsity for long-context modeling. Inspired by these advances, MTS representation learning has adopted masking to mitigate redundancy. Methods like Ti-MAE~\cite{li2023timaeselfsupervisedmaskedtime} use masked autoencoding for forecasting alignment, PrimeNet~\cite{Chowdhury_Li_Zhang_Hong_Gupta_Shang_2023} implements density-aware masking for irregular sampling, and TS-MVP~\cite{ZhongTS-MVPADMA} employs probability-based masking for multi-view alignment. Others including STCR~\cite{Lee24Access} and TS2Vec~\cite{Yue_Wang_Duan_Yang_Huang_Tong_Xu_2022} incorporate random masking in contrastive frameworks. However, these MTS methods predominantly rely on static or predetermined masking policies that lack adaptability to co-evolve with learning, hindering dynamic suppression of evolving redundant patterns or prioritization of cluster-salient features.

\section{Proposed Method}

Existing masking strategies are limited by their inability to adapt to MTS clustering tasks. To address this, this paper proposes \textbf{EMTC} with two core components: \textbf{IVM} and \textbf{MEV} modules. As illustrated in Figure~\ref{Overview}, learning of the model involves two pathways:
\begin{itemize}
\item \textbf{CRL}: Dynamically suppresses temporal redundancy while learning robust multi-view representations.
\item \textbf{CMC}: Structures embedding space under cluster-aware contrastive objectives to enhance cluster separability.
\end{itemize}

Let $X_i = [\mathbf{x}_1, \mathbf{x}_2, \ldots, \mathbf{x}_T] \in \mathbb{R}^{T \times D}$ denotes an MTS sample of length $T$ with $D$ variates, where each $\mathbf{x}_t \in \mathbb{R}^D$ represents the multivariate observation at timestamp $t$. Given a dataset $X = \{X_i\}_{i=1}^N$, the objective is to partition the $N$ samples into $g$ disjoint clusters $\mathcal{C} = \{\mathcal{C}_1, \mathcal{C}_2, \ldots, \mathcal{C}_g\}$.

\subsection{IVM: Importance-aware Variate-wise Masking}
\label{subsec:IVM}
To explicitly facilitate dynamic redundancy suppression, learnable IVM is designed to adapt the masking to the MTS clustering tasks. The clustering-friendly evolving masking is realized in three stages: single-variate view generation, content-aware importance evaluation, and redundant timestamp masking.

\textbf{Single-Variate View Generation:} To facilitate content-aware importance evaluation, we first generate single-variate view embeddings from the original input $X$:
\begin{equation}
Z^{(d)} = f_{\theta}^{(d)}\left(X\right), \quad \forall d \in \{1,2,\ldots,D\},
\label{single_view_generation}
\end{equation}
where $Z^{(d)}$ provides the single-variate embedding that preserves the integrity of individual variate information.

\textbf{Content-Aware Importance Evaluation:} Based on the single-variate embeddings, the importance of each timestamp of sample $Z_{i}^{(d)}$ is evaluated by content-aware attention:
\begin{equation}
S_i^{(d)} = \text{Softmax}\left(\frac{Q_i^{(d)}(K_i^{(d)})^T}{\sqrt{d_k}}\right)Z_i^{(d)}, \label{importance_evaluation}
\end{equation}
where the attention mechanism optimized with the clustering objective can dynamically refine its focus on discriminative timestamps to enhance inter-cluster separation. Unlike conventional attention-weighted MTS that are often misled by highly activated yet non-informative patterns, our attention-based masking directly excludes the timestamps that are relatively clustering-irrelevant, ensuring dynamic alignment with learned cluster semantics, achieving precise and task-aware MTS redundancy suppression with the CRL and CMC dual-learning pathways.

\textbf{Redundant Timestamp Masking:} With the attention-based importance evaluation, thresholding is employed to mask less-important timestamps:
\begin{equation}
M^{(d)}_i(t) = \begin{cases} 
1 & \text{if } S_i^{(d)}(t) \geq \epsilon \\
0 & \text{otherwise}
\end{cases}, \quad \forall t \in [1,T],
\label{get_mask}
\end{equation}
where $\epsilon$ is a predefined threshold to filter out redundant timestamps. $M^{(d)}_i(t)$ denotes the binary mask value at timestamp $t$. The computed mask set $\mathcal{M}_i = \{M_i^{(1)}, M_i^{(2)}, \ldots, M_i^{(D)}\} \in \mathbb{R}^{T \times D}$ is then applied to $D$ variates of the $i$-th original sample $X_i$ through element-wise multiplication $X \odot \mathcal{M}$ to obtain the masked input $\widetilde{X}$ in the next epoch. In the first learning epoch, a randomly initialized mask set $\mathcal{M} \in \mathbb{R}^{N \times T \times D}$ is adopted to obtain $\widetilde{X}$.

\subsection{MEV: Multi-Endogenous Views Generation}
\label{subsec:MEV}
To overcome the limitations of single-view representations in capturing complex multivariate patterns, this paper proposes to perform MEV generation to obtain complementary MTS perspectives. That is, learning upon the multi-view MTS ensures robust representation learning while providing natural regularization for the evolving masking. Specifically, the masked input $\widetilde{X}$ obtained from IVM is processed through MEV embeddings generation:
\begin{equation}
F^{(v)} = f_{\Theta}^{(v)}\left(\widetilde{X}\right), \quad \forall v \in \{1,2,\ldots,V\},
\label{MEV_Generate}
\end{equation}
where the $F^{(v)}$ denotes the $v$-th endogenous view. Such an MEV design ensures comprehensive pattern capture while maintaining the integrity of view-specific information for masking decisions.

With the endogenous views obtained based on the IVM masking, the model performs a dual-path representation learning, comprising CRL for IVM consolidation and CMC for cluster discriminative representation enhancement, which are introduced below.

\subsection{CRL: Consistency and Reconstruction Learning}
\label{subsec:cross_views_consistency}

CRL involves two complementary mechanisms: intra-view reconstruction learning helps the MEV retain the semantic structure of the original time-series, while inter-view reconstruction learning establishes semantic bridges between different views. For each view $F^{(v)}$, the model reconstructs the original MTS $X$ through $X^{(v)} = R^{(v)}(F^{(v)})$, where the decoder $R^{(v)}$ maps the endogenous view embedding $F^{(v)}$ to the reconstructed sequence $X^{(v)}$. The intra-view reconstruction learning can then be formulated as:
\begin{equation}
\mathcal{L}_{\text{intra}} = \sum_{v = 1}^{V}\left\| X - 
X^{(v)} \right\|_{F}^{2}. \label{intra_1}
\end{equation}

Such self-reconstruction acts as a regularizer to retain the semantic structure during modal training and preserve discriminative temporal features for clustering. 

By contrast, inter-view reconstruction establishes inter-view semantic consistency to enhance representation robustness by measuring the discrepancy between the original and transformed embeddings across views:
\begin{equation}
\mathcal{L}_{\text{inter}} = \sum_{i = 1}^{V}\sum_{j = i + 1}^{V}\mathcal{L}_{\text{inter}}^{i \rightarrow j} + \mathcal{L}_{\text{inter}}^{j \rightarrow i}, \label{inter_1}
\end{equation}
specifically, for each pair of endogenous view embeddings $(F^{(i)},F^{(j)})$, the loss between $F^{(j)}$ and the transformed embedding $F^{(i \rightarrow j)} = \mathcal{H}_{i \rightarrow j}(F^{(i)})$ is computed, where $\mathcal{H}_{i \rightarrow j}$ is a transformation decoder that maps embeddings from view $F^{(i)}$ to view $F^{(j)}$ as:

\begin{equation}
\mathcal{L}_{\text{inter}}^{i \rightarrow j} = \left\| F^{(j)} - F^{(i \rightarrow j)} \right\|_{F}^{2}.\label{inter_2}
\end{equation}

CRL simultaneously considers single-view fidelity and multi-view consistency to learn representations that are both consistent with the key features of the masked MTS and semantically coherent across endogenous view embeddings. More importantly, the reconstruction learning and consistency learning based on MEV mitigate the risk of overfitting to variate-specific knowledge and losing critical single temporal information caused by the crisp IVM masking. 

\subsection{CMC: Clustering-Guided MEV Contrastive Learning}
\label{subsec:contrastive_learning}
To explicitly structure the embedding space for cluster separation, CMC is also introduced to leverage cluster assignments as supervisory signals. The process begins by aggregating the complementary information from all MEV embeddings. Specifically, temporal pooling is applied to each endogenous view embedding $F^{(v)}$, followed by an MEV fusion operation, yielding a fused representation:
\begin{equation}
F^{(\text{fused})} = F^{(1)} \circledast F^{(2)} \circledast \cdots \circledast F^{(V)},
\label{MEV_Fusion}
\end{equation}
where $\circledast$ denotes the fusion operation that integrates information across all views. The integrated representation $F^{(\text{fused})}$ combines discriminative patterns from different views, forming a robust basis for clustering. For simplicity but without losing generality, $F^{(\text{fused})}$ is denoted as $F$ hereinafter. Subsequently, cluster labels $\mathcal{C}$ are obtained by performing $k$-means clustering on $F$ at each training epoch:
\begin{equation}
\mathcal{C} = \text{Cluster}(F). \label{clustering}
\end{equation}

These dynamically updated cluster labels guide the construction of pairs for contrastive learning. For an anchor sample $F_{i}$, samples sharing the same cluster label form positive pairs, with all the positive samples denoted as $\mathcal{P}(i^+)$. The other samples outside the cluster of $F_{i}$ serve as negative pairs, which can be written as $\mathcal{N}(i^-)$.

The discriminative contrastive objective, adapted from the foundation laid in~\cite{CCGC2023,PengAAAI2024}, is utilized to maximize the cosine similarity within each cluster (positive pairs) and enforce robust separation between distinct clusters (negative pairs), which can be expressed as:
\begin{equation}
\mathcal{L}_{\text{contra}} = -\sum_{i=1}^{N} \log \frac{L_i^+}{L_i^+ + L_i^-}.
\label{conta_1}
\end{equation}
$L_i^+$ is the loss contributed by the positive pairs w.r.t. $F_i$, which can be written as:
\begin{equation}
L_i^+=\sum_{\mathcal{P}(i^+)} \exp \left( \frac{\text{sim} (i, i^{+})}{\tau} \right),
\label{contra_pos}
\end{equation}
where $\tau$ serves as a temperature parameter controlling the sharpness of the similarity distribution, and $\text{sim}\left(i, i^{+}\right)$ denotes the cosine similarity between positive sample pairs, which can be instantiated as:
\begin{equation}
\text{sim} (i, i^{+}) = \frac{F^\top_{i}  F_{i^+}}{\left\| 
F_{i} \right\| \left\| F_{i^+} \right\|}\ \ \text{s.t.}\ \ F_{i^+}\in\mathcal{P}(i^+). \label{conta_2}
\end{equation}
The definitions for negative pairs, i.e., $L_i^-$ and $\text{sim} (i, i^{-})$, are analogous.

Such a dynamic clustering-guided representation learning strategy ensures a close connection between the contrastive learning process and the clustering objective. By continuously updating the cluster labels throughout the training, the model can effectively refine the MTS representation to fit newly learned cluster distributions.

\begin{algorithm}[t]
\caption{EMTC: Evolving-masked MTS Clustering.}
\label{alg:emtc}
\textbf{Input}: MTS dataset $X$, cluster number $g$.
\begin{algorithmic}[1]
\State Initialize IVM and MEV components.
\Repeat
    \State IVM: Update mask set via Eqs.~(\ref{single_view_generation}-\ref{mask_compare});
    \State MEV: Generate embedding $F^{(v)}$ via Eq.~(\ref{MEV_Generate});
    \State CRL: Compute $\mathcal{L}_{\text{intra}}$ and $\mathcal{L}_{\text{inter}}$ via Eqs.~(\ref{intra_1}-\ref{inter_2});
    \State Update cluster labels $\mathcal{C}$ via Eq.~(\ref{clustering});
    \State CMC: Compute $\mathcal{L}_{\text{contra}}$ via Eqs.~(\ref{conta_1}) and (\ref{conta_2});
    \State Compute total loss $\mathcal{L}_{\text{total}}$ for Adam optimization;
\Until{convergence} 
\end{algorithmic}
\textbf{Output}: Final MTS clustering results $\mathcal{C}$.
\end{algorithm}

\subsection{Model Training and Clustering}
\label{subsec:total_loss}

The overall loss $\mathcal{L}_{\text{total}}$ with previously introduced loss terms integrated through balancing coefficients, i.e., $\alpha$ for $\mathcal{L}_{\text{intra}}$ and $\beta$ for $\mathcal{L}_{\text{inter}}$, can be written as:
\begin{equation}
\mathcal{L}_{\text{total}} = \mathcal{L}_{\text{contra}} + \alpha \mathcal{L}_{\text{intra}} + \beta \mathcal{L}_{\text{inter}}. \label{eq:13}
\end{equation}

The model parameters are optimized using the Adam optimizer.
The full EMTC algorithm for MTS representation learning and clustering is summarized as Algorithm~\ref{alg:emtc}.

\section{Experiments}

Eight types of experiments have been designed to comprehensively evaluate the proposed EMTC: 1) \textbf{Clustering Performance Comparison} with eight SOTA methods on 15 datasets; 2) \textbf{Significance Test} using BD-test with 95\% and 99\% confidence intervals; 3) \textbf{Key Component Ablation} validating IVM and MEV modules; 4) \textbf{Loss Ablation} validating each loss term; 5) \textbf{Masking Strategies Comparison} illustrating the rationality of our IVM design; 6) \textbf{Convergence Analysis} tracking loss dynamics; 7) \textbf{Efficiency Evaluation} measuring computational requirements and scalability; 8) \textbf{Cluster Effect Visualization} using t-SNE for qualitative assessment.

\textbf{15 real-world benchmark datasets} from the UEA MTS archive~\cite{bagnall2018uea}, spanning diverse 
domains with varying number of samples, sequence lengths and feature dimensions. Detailed statistics are presented in Table~\ref{tbl_summary}. \textbf{Four standard metrics} are employed: Accuracy (ACC), F1-Score (F1), Normalized Mutual Information (NMI) and Adjusted Rand Index (ARI). \textbf{Eight SOTA counterparts} are compared, including deep MTS clustering methods FEI~\cite{Fu_Hu_2025}, FCACC~\cite{wang2025fcacc1}, TimesURL~\cite{Liu_Chen_2024}, UNITS~\cite{NEURIPS2024_fe248e22}, USLA~\cite{Zhang23MTSC}, Ti-MAE~\cite{li2023timaeselfsupervisedmaskedtime}, MHCCL~\cite{mhcclaaai}, and conventional feature engineering method T-GMRF~\cite{TKDEDing2023}. All counterparts are tuned following their original papers, with parameters optimized for each dataset individually. Our architecture utilizes dilated convolution for encoders $f_{\Theta}^{(v)}$ and $f_{\theta}^{(d)}$, while employing MLP decoder with specific configurations: six-layer networks for intra-view reconstruction $(R^{(v)})$ and four-layer networks for inter-view transformation $(\mathcal{H}_{i \rightarrow j})$. The fused representation $F^{(\text{fused})}$ is obtained through the fusion operation $\circledast$ that averages the sum of all $F^{(v)}$, which can be extended to weighted combinations. We set the batch size to 64 and train for a maximum of 200 epochs. All experiments are implemented in PyTorch 1.8.0 on a NVIDIA RTX4090 GPU, 20GB RAM.

\begin{table}[!t]
\centering
\resizebox{\linewidth}{!}{\begin{tabular}{c|c|ccc|c}
 \toprule
            No. & Datasets & $N$ & $T$ & $D$ & $g$ \\ 
            \midrule
            1 &  BasicMotions  & 40 & 100 & 6 & 4 \\ 
            2 &  Cricket  & 72 & 1197 & 6 & 12 \\ 
            3 &  DuckDuckGeese  & 40 & 270 & 1345 & 5 \\ 
            4 &  EigenWorms & 131 & 17984 & 6 & 5 \\ 
            5 &  Epilepsy & 138 & 206 & 3 & 4 \\ 
            6 &  FingerMovements & 100 & 50 & 28 & 2 \\ 
            7 &  HandMovementDirection & 147 & 400 & 10 & 4 \\ 
            8 &  Heartbeat  & 205 & 405 & 61 & 2 \\ 
            9 &  MotorImagery  & 100 & 3000 & 64 & 2 \\ 
            10 &  NATOPS  & 180 & 51 & 24 & 6 \\ 
            11 &  PEMS-SF  & 173 & 144 & 963 & 7 \\ 
            12 &  RacketSports  & 152 & 30 & 6 & 4 \\ 
            13 &  SelfRegulationSCP1 & 293 & 896 & 6 & 2 \\ 
            14 &  SelfRegulationSCP2  & 180 & 1152 & 7 & 2 \\
            15 &  StandWalkJump  & 15 & 2500 & 4 & 3 \\ 
            \bottomrule
\end{tabular}}
\caption{Statistics of datasets. $g$ indicates the `true' number of clusters provided by the labels of the datasets.}
\label{tbl_summary}
\end{table}

\definecolor{bestblue}{HTML}{FFDDDD}    
\definecolor{secondblue}{HTML}{DEEBF7}  

\begin{table*}[!t]
\centering

\resizebox{\textwidth}{!}{
\begin{tabular}{l|c|ccccccccc}
\toprule
\multirow{2}{*}{Datasets} & \multirow{2}{*}{Metrics} & \multicolumn{9}{c}{Methods} \\
\cmidrule(lr){3-11}
 & & \textbf{EMTC (ours)} & FEI (AAAI'25) & FCACC (arxiv'25) & TimesURL (AAAI'24) & UNITS (NeurIPS'24) & USLA (TPAMI'23) & Ti-MAE (arxiv'23) & MHCCL (AAAI'23) & T-GMRF (TKDE'23) \\
\midrule
\multirow{4}{*}{BasicMotions} & ACC & \cellcolor{secondblue}0.9083 ± 0.0656 & 0.4667 ± 0.0589 & 0.7250 ± 0.0000 & 0.6917 ± 0.0118 & 0.7667 ± 0.1662 & 0.8750 ± 0.0000 & \cellcolor{bestblue}0.9750 ± 0.0000 & 0.1750 ± 0.0354 & 0.3250 ± 0.0000 \\
 & F1 & \cellcolor{secondblue}0.9057 ± 0.0675 & 0.4975 ± 0.1143 & 0.8713 ± 0.0000 & 0.8446 ± 0.0103 & 0.8665 ± 0.0961 & 0.3801 ± 0.1374 & \cellcolor{bestblue}0.9749 ± 0.0000 & 0.2002 ± 0.0311 & 0.2375 ± 0.0000 \\
 & NMI & 0.8391 ± 0.1189 & 0.4555 ± 0.0996 & 0.8294 ± 0.0000 & 0.8150 ± 0.0045 & \cellcolor{secondblue}0.8515 ± 0.1069 & 0.8064 ± 0.0000 & \cellcolor{bestblue}0.9404 ± 0.0000 & 0.6895 ± 0.0000 & 0.1623 ± 0.0000 \\
 & ARI & \cellcolor{secondblue}0.7910 ± 0.1490 & 0.2258 ± 0.1205 & 0.6754 ± 0.0000 & 0.6337 ± 0.0088 & 0.7310 ± 0.1904 & 0.7218 ± 0.0000 & \cellcolor{bestblue}0.9315 ± 0.0000 & 0.5152 ± 0.0000 & 0.0092 ± 0.0000 \\

\midrule

\multirow{4}{*}{Cricket} & ACC & \cellcolor{bestblue}0.5972 ± 0.0227 & 0.4537 ± 0.0755 & 0.4259 ± 0.0429 & 0.4722 ± 0.0196 & 0.3287 ± 0.0236 & 0.3611 ± 0.0000 & \cellcolor{secondblue}0.4832 ± 0.0226 & 0.0648 ± 0.0398 & 0.1389 ± 0.0000 \\
 & F1 & \cellcolor{bestblue}0.6317 ± 0.0366 & \cellcolor{secondblue}0.5136 ± 0.0501 & 0.4301 ± 0.0392 & 0.4913 ± 0.0134 & 0.3556 ± 0.0187 & 0.0573 ± 0.0367 & 0.4903 ± 0.0124 & 0.0600 ± 0.0321 & 0.1102 ± 0.0000 \\
 & NMI & \cellcolor{secondblue}0.6574 ± 0.0273 & 0.6348 ± 0.0484 & 0.5360 ± 0.0400 & 0.6059 ± 0.0027 & 0.4115 ± 0.0199 & 0.4701 ± 0.0025 & 0.6059 ± 0.0027 & \cellcolor{bestblue}0.7264 ± 0.0077 & 0.3218 ± 0.0105 \\
 & ARI & \cellcolor{secondblue}0.3436 ± 0.0303 & 0.2866 ± 0.0817 & 0.2413 ± 0.0551 & 0.2638 ± 0.0243 & 0.0473 ± 0.0191 & 0.0943 ± 0.0020 & 0.2633 ± 0.0143 & \cellcolor{bestblue}0.4059 ± 0.0096 & 0.0138 ± 0.0016 \\

\midrule

\multirow{4}{*}{DuckDuckGeese} & ACC & \cellcolor{bestblue}0.4800 ± 0.0163 & 0.2933 ± 0.0249 & 0.3550 ± 0.0087 & 0.3133 ± 0.0340 & 0.3667 ± 0.0094 & 0.3200 ± 0.0000 & \cellcolor{secondblue}0.3733 ± 0.0189 & 0.1933 ± 0.0249 & 0.2600 ± 0.0000 \\
 & F1 & \cellcolor{bestblue}0.4917 ± 0.0357 & 0.2445 ± 0.0185 & 0.3300 ± 0.0070 & 0.3065 ± 0.0294 & \cellcolor{secondblue}0.3980 ± 0.0095 & 0.1457 ± 0.0169 & 0.3602 ± 0.0224 & 0.1447 ± 0.0241 & 0.1758 ± 0.0000 \\
 & NMI & \cellcolor{bestblue}0.2719 ± 0.0340 & 0.1622 ± 0.0028 & 0.1099 ± 0.0054 & 0.1296 ± 0.0616 & 0.1759 ± 0.0444 & 0.0970 ± 0.0000 & \cellcolor{secondblue}0.2261 ± 0.0330 & 0.0915 ± 0.0000 & 0.1981 ± 0.0000 \\
 & ARI & \cellcolor{bestblue}0.1239 ± 0.0350 & 0.0041 ± 0.0059 & 0.0275 ± 0.0047 & -0.0071 ± 0.0301 & 0.0368 ± 0.0142 & -0.0117 ± 0.0000 & \cellcolor{secondblue}0.0374 ± 0.0214 & -0.0209 ± 0.0000 & 0.0070 ± 0.0000 \\

\midrule

\multirow{4}{*}{EigenWorms} & ACC & \cellcolor{bestblue}0.4707 ± 0.0095 & 0.4580 ± 0.0125 & 0.4530 ± 0.0125 & \cellcolor{secondblue}0.4633 ± 0.0134 & 0.3410 ± 0.0401 & 0.3486 ± 0.0036 & 0.4540 ± 0.0221 & 0.3486 ± 0.0036 & 0.4540 ± 0.0225 \\
 & F1 & \cellcolor{bestblue}0.3713 ± 0.0549 & 0.3404 ± 0.0779 & 0.3333 ± 0.0224 & 0.3204 ± 0.0789 & 0.2894 ± 0.0168 & 0.2245 ± 0.0625 & 0.3401 ± 0.0023 & 0.2245 ± 0.0625 & \cellcolor{secondblue}0.3404 ± 0.0023 \\
& NMI & \cellcolor{bestblue}0.1538 ± 0.0275 & 0.1275 ± 0.0333 & 0.1221 ± 0.0313 & 0.1275 ± 0.0332 & 0.0859 ± 0.0256 & \cellcolor{secondblue}0.1445 ± 0.0030 & 0.1133 ± 0.0133 & \cellcolor{secondblue}0.1445 ± 0.0030 & 0.1175 ± 0.0133 \\
& ARI & \cellcolor{bestblue}0.1012 ± 0.0435 & 0.1010 ± 0.0627 & \cellcolor{secondblue}0.1011 ± 0.0227 & 0.1000 ± 0.0627 & 0.0432 ± 0.0394 & 0.0760 ± 0.0017 & 0.1011 ± 0.0627 & 0.0760 ± 0.0017 & 0.1000 ± 0.0627 \\

\midrule

\multirow{4}{*}{Epilepsy} & ACC & \cellcolor{bestblue}0.5556 ± 0.0239 & 0.4082 ± 0.0208 & 0.5236 ± 0.0107 & 0.4517 ± 0.0034 & 0.4179 ± 0.0304 & \cellcolor{secondblue}0.5338 ± 0.0034 & 0.5145 ± 0.0000 & 0.2053 ± 0.0754 & 0.4928 ± 0.0000 \\
 & F1 & \cellcolor{bestblue}0.5519 ± 0.0276 & 0.4077 ± 0.0065 & 0.5193 ± 0.0093 & \cellcolor{secondblue}0.5226 ± 0.0048 & 0.4116 ± 0.0414 & 0.1491 ± 0.0204 & 0.5133 ± 0.0001 & 0.2017 ± 0.0712 & 0.3467 ± 0.0000 \\
  & NMI & 0.2783 ± 0.0182 & 0.1851 ± 0.0462 & \cellcolor{secondblue}0.2834 ± 0.0152 & 0.1882 ± 0.0170 & 0.1370 ± 0.0371 & 0.2576 ± 0.0008 & 0.2748 ± 0.0005 & 0.2387 ± 0.0038 & \cellcolor{bestblue}0.4436 ± 0.0000 \\
 & ARI & 0.2068 ± 0.0273 & 0.1035 ± 0.0435 & \cellcolor{secondblue}0.2165 ± 0.0115 & 0.1234 ± 0.0078 & 0.0722 ± 0.0346 & 0.1419 ± 0.0020 & 0.1950 ± 0.0001 & 0.2033 ± 0.0030 & \cellcolor{bestblue}0.2707 ± 0.0000 \\

\midrule

\multirow{4}{*}{\shortstack[l]{Finger\\Movements}} & ACC & \cellcolor{bestblue}0.5967 ± 0.0094 & 0.5100 ± 0.0141 & 0.5000 ± 0.0000 & 0.5100 ± 0.0000 & 0.5800 ± 0.0535 & 0.5000 ± 0.0000 & \cellcolor{secondblue}0.5800 ± 0.0000 & 0.5133 ± 0.0377 & 0.5200 ± 0.0163 \\
 & F1 & \cellcolor{bestblue}0.5892 ± 0.0151 & 0.5071 ± 0.0146 & 0.3503 ± 0.0000 & 0.4954 ± 0.0000 & \cellcolor{secondblue}0.5771 ± 0.0514 & 0.4770 ± 0.0001 & 0.5758 ± 0.0000 & 0.4867 ± 0.0422 & 0.4028 ± 0.0679 \\
& NMI & \cellcolor{bestblue}0.0320 ± 0.0030 & 0.0010 ± 0.0014 & 0.0181 ± 0.0000 & 0.0002 ± 0.0000 & \cellcolor{secondblue}0.0311 ± 0.0284 & 0.0001 ± 0.0000 & 0.0185 ± 0.0000 & 0.0080 ± 0.0000 & 0.0272 ± 0.0117 \\
 & ARI & \cellcolor{bestblue}0.0287 ± 0.0071 & -0.0087 ± 0.0017 & -0.0006 ± 0.0000 & -0.0088 ± 0.0000 & \cellcolor{secondblue}0.0276 ± 0.0326 & -0.0085 ± 0.0000 & 0.0160 ± 0.0000 & -0.0017 ± 0.0000 & -0.0001 ± 0.0007 \\

\midrule

\multirow{4}{*}{\shortstack[l]{HandMovement\\ Direction}} & ACC & \cellcolor{secondblue}0.4640 ± 0.0064 & 0.3333 ± 0.0169 & \cellcolor{bestblue}0.4696 ± 0.0112 & 0.3514 ± 0.0191 & 0.3514 ± 0.0110 & 0.3649 ± 0.0000 & 0.3243 ± 0.0000 & 0.2523 ± 0.0337 & 0.4189 ± 0.0000 \\
 & F1 & \cellcolor{secondblue}0.3737 ± 0.0262 & 0.3071 ± 0.0601 & \cellcolor{bestblue}0.3993 ± 0.0192 & 0.3434 ± 0.0194 & 0.3222 ± 0.0164 & 0.2717 ± 0.0475 & 0.3186 ± 0.0047 & 0.2293 ± 0.0502 & 0.2661 ± 0.0000 \\
  & NMI & 0.0914 ± 0.0137 & 0.0454 ± 0.0124 & \cellcolor{secondblue}0.0961 ± 0.0118 & 0.0652 ± 0.0207 & 0.0378 ± 0.0107 & 0.0883 ± 0.0004 & 0.0527 ± 0.0004 & 0.0874 ± 0.0000 & \cellcolor{bestblue}0.1338 ± 0.0000 \\
 & ARI & \cellcolor{secondblue}0.0767 ± 0.0068 & -0.0010 ± 0.0187 & \cellcolor{bestblue}0.0994 ± 0.0193 & -0.0039 ± 0.0068 & -0.0076 ± 0.0079 & 0.0112 ± 0.0001 & -0.0130 ± 0.0007 & 0.0031 ± 0.0000 & 0.0245 ± 0.0000 \\

\midrule

\multirow{4}{*}{Heartbeat} & ACC & \cellcolor{bestblue}0.7463 ± 0.0040 & \cellcolor{secondblue}0.7138 ± 0.0023 & 0.5646 ± 0.0040 & 0.7122 ± 0.0000 & 0.7122 ± 0.0000 & \cellcolor{secondblue}0.7138 ± 0.0023 & 0.7122 ± 0.0000 & 0.4341 ± 0.0000 & 0.6911 ± 0.0506 \\
 & F1 & 0.6134 ± 0.0329 & \cellcolor{bestblue}0.8330 ± 0.0016 & 0.5549 ± 0.0051 & \cellcolor{secondblue}0.8319 ± 0.0000 & \cellcolor{secondblue}0.8319 ± 0.0000 & 0.6014 ± 0.0011 & 0.4160 ± 0.0000 & 0.4325 ± 0.0000 & 0.5952 ± 0.0303 \\
 & NMI & \cellcolor{bestblue}0.0733 ± 0.0123 & 0.0158 ± 0.0028 & 0.0402 ± 0.0074 & 0.0178 ± 0.0000 & 0.0178 ± 0.0000 & 0.0083 ± 0.0022 & 0.0099 ± 0.0000 & 0.0004 ± 0.0000 & \cellcolor{secondblue}0.0578 ± 0.0160 \\
 & ARI & \cellcolor{bestblue}0.1531 ± 0.0273 & -0.0098 ± 0.0027 & \cellcolor{secondblue}0.0105 ± 0.0014 & -0.0117 ± 0.0000 & -0.0117 ± 0.0000 & -0.0098 ± 0.0027 & -0.0117 ± 0.0000 & -0.0087 ± 0.0000 & -0.0159 ± 0.0445 \\

\midrule

\multirow{4}{*}{MotorImagery} & ACC & \cellcolor{bestblue}0.6500 ± 0.0216 & 0.5100 ± 0.0000 & \cellcolor{secondblue}0.5700 ± 0.0122 & 0.5200 ± 0.0000 & 0.5500 ± 0.0327 & 0.5100 ± 0.0000 & 0.5100 ± 0.0000 & 0.5233 ± 0.0660 & 0.5500 ± 0.0141 \\
 & F1 & \cellcolor{bestblue}0.6452 ± 0.0213 & 0.3924 ± 0.0263 & \cellcolor{secondblue}0.5572 ± 0.0232 & 0.3912 ± 0.0000 & 0.5428 ± 0.0391 & 0.3552 ± 0.0000 & 0.3552 ± 0.0000 & 0.5175 ± 0.0668 & 0.4812 ± 0.0499 \\
 & NMI & \cellcolor{bestblue}0.0737 ± 0.0229 & 0.0127 ± 0.0161 & 0.0168 ± 0.0125 & 0.0159 ± 0.0000 & 0.0107 ± 0.0102 & 0.0186 ± 0.0000 & 0.0186 ± 0.0000 & 0.0152 ± 0.0000 & \cellcolor{secondblue}0.0220 ± 0.0037 \\
 & ARI & \cellcolor{bestblue}0.0831 ± 0.0274 & -0.0019 ± 0.0014 & \cellcolor{secondblue}0.0152 ± 0.0014 & 0.0000 ± 0.0000 & 0.0048 ± 0.0132 & 0.0000 ± 0.0000 & 0.0000 ± 0.0000 & 0.0101 ± 0.0000 & 0.0059 ± 0.0030 \\

\midrule

\multirow{4}{*}{NATOPS} & ACC & \cellcolor{bestblue}0.6185 ± 0.0094 & 0.4593 ± 0.0723 & \cellcolor{secondblue}0.6056 ± 0.0000 & 0.4167 ± 0.1179 & 0.3833 ± 0.0552 & 0.3944 ± 0.0000 & 0.5278 ± 0.0000 & 0.0815 ± 0.0498 & 0.2167 ± 0.0045 \\
 & F1 & \cellcolor{bestblue}0.6205 ± 0.0130 & 0.5502 ± 0.0698 & \cellcolor{secondblue}0.5821 ± 0.0000 & 0.4102 ± 0.1252 & 0.4029 ± 0.0367 & 0.1579 ± 0.0518 & 0.5294 ± 0.0000 & 0.0943 ± 0.0607 & 0.1316 ± 0.0045 \\
 & NMI & 0.5503 ± 0.0058 & 0.5450 ± 0.0526 & \cellcolor{secondblue}0.7277 ± 0.0000 & 0.2961 ± 0.1807 & 0.3045 ± 0.0527 & 0.3719 ± 0.0000 & 0.5243 ± 0.0000 & \cellcolor{bestblue}0.7606 ± 0.0011 & 0.0524 ± 0.0005 \\
 & ARI & 0.3966 ± 0.0155 & 0.3459 ± 0.0662 & \cellcolor{secondblue}0.5560 ± 0.0000 & 0.1696 ± 0.1477 & 0.1520 ± 0.0494 & 0.1846 ± 0.0000 & 0.3487 ± 0.0000 & \cellcolor{bestblue}0.5889 ± 0.0028 & 0.0027 ± 0.0005 \\

\midrule

\multirow{4}{*}{PEMS-SF} & ACC & \cellcolor{secondblue}0.5780 ± 0.0164 & 0.4355 ± 0.0196 & 0.4697 ± 0.0085 & \cellcolor{bestblue}0.6012 ± 0.0000 & 0.5453 ± 0.0607 & 0.2370 ± 0.0000 & 0.4682 ± 0.0000 & 0.1715 ± 0.0027 & 0.2004 ± 0.0027 \\
 & F1 & 0.6215 ± 0.0398 & 0.4310 ± 0.0260 & 0.5328 ± 0.0085 & \cellcolor{bestblue}0.6419 ± 0.0000 & \cellcolor{secondblue}0.6408 ± 0.0433 & 0.1118 ± 0.0186 & 0.4506 ± 0.0021 & 0.1313 ± 0.0111 & 0.1150 ± 0.0052 \\
 & NMI & 0.5316 ± 0.0347 & 0.3722 ± 0.0160 & 0.4821 ± 0.0068 & \cellcolor{bestblue}0.5824 ± 0.0000 & \cellcolor{secondblue}0.5684 ± 0.0233 & 0.0819 ± 0.0022 & 0.5275 ± 0.0009 & 0.2572 ± 0.0003 & 0.0578 ± 0.0007 \\
 & ARI & 0.3611 ± 0.0277 & 0.2164 ± 0.0249 & 0.3028 ± 0.0069 & \cellcolor{bestblue}0.4517 ± 0.0000 & \cellcolor{secondblue}0.3739 ± 0.0555 & 0.0147 ± 0.0019 & 0.3197 ± 0.0018 & 0.0814 ± 0.0004 & 0.0007 ± 0.0006 \\

\midrule

\multirow{4}{*}{RacketSports} & ACC & \cellcolor{bestblue}0.4715 ± 0.0135 & 0.4254 ± 0.0265 & 0.3766 ± 0.0028 & 0.3750 ± 0.0000 & 0.3772 ± 0.0499 & 0.3553 ± 0.0000 & \cellcolor{secondblue}0.4276 ± 0.0000 & 0.2851 ± 0.0967 & 0.2895 ± 0.0000 \\
 & F1 & \cellcolor{bestblue}0.4657 ± 0.0695 & 0.3949 ± 0.0301 & 0.3757 ± 0.0034 & 0.3489 ± 0.0000 & 0.3748 ± 0.0465 & 0.3078 ± 0.0467 & \cellcolor{secondblue}0.4347 ± 0.0001 & 0.2911 ± 0.0972 & 0.1396 ± 0.0000 \\
 & NMI & \cellcolor{secondblue}0.2004 ± 0.0520 & 0.1310 ± 0.0264 & 0.1248 ± 0.0100 & 0.0913 ± 0.0000 & 0.0645 ± 0.0408 & 0.0662 ± 0.0000 & \cellcolor{bestblue}0.2178 ± 0.0009 & 0.1629 ± 0.0231 & 0.0359 ± 0.0000 \\
 & ARI & \cellcolor{secondblue}0.1235 ± 0.0372 & 0.0986 ± 0.0374 & 0.0722 ± 0.0048 & 0.0449 ± 0.0000 & 0.0349 ± 0.0314 & 0.0392 ± 0.0000 & \cellcolor{bestblue}0.1293 ± 0.0012 & 0.1196 ± 0.0107 & -0.0006 ± 0.0000 \\

\midrule

\multirow{4}{*}{\shortstack[l]{SelfRegulation\\SCP1}} & ACC & \cellcolor{secondblue}0.8510 ± 0.0126 & 0.8385 ± 0.0621 & 0.5742 ± 0.0044 & 0.7247 ± 0.0903 & 0.7270 ± 0.1401 & 0.6007 ± 0.0000 & \cellcolor{bestblue}0.8840 ± 0.0000 & 0.4790 ± 0.0595 & 0.5040 ± 0.0016 \\
 & F1 & \cellcolor{secondblue}0.8500 ± 0.0135 & 0.8337 ± 0.0687 & 0.5038 ± 0.0098 & 0.6979 ± 0.1211 & 0.7234 ± 0.1418 & 0.3154 ± 0.0000 & \cellcolor{bestblue}0.8838 ± 0.0000 & 0.4671 ± 0.0611 & 0.3415 ± 0.0018 \\
 & NMI & 0.4146 ± 0.0207 & \cellcolor{secondblue}0.4177 ± 0.0873 & 0.0537 ± 0.0031 & 0.2403 ± 0.0974 & 0.2489 ± 0.1753 & 0.0790 ± 0.0000 & \cellcolor{bestblue}0.4878 ± 0.0000 & 0.0133 ± 0.0000 & 0.0189 ± 0.0001 \\
 & ARI & \cellcolor{secondblue}0.4916 ± 0.0355 & 0.4720 ± 0.1577 & 0.0214 ± 0.0026 & 0.2327 ± 0.1393 & 0.2822 ± 0.2000 & 0.0389 ± 0.0000 & \cellcolor{bestblue}0.5883 ± 0.0000 & 0.0129 ± 0.0000 & 0.0000 ± 0.0000 \\
\midrule

\multirow{4}{*}{\shortstack[l]{SelfRegulation\\SCP2}} & ACC & \cellcolor{bestblue}0.6000 ± 0.0253 & 0.5630 ± 0.0114 & \cellcolor{secondblue}0.5972 ± 0.0028 & 0.5222 ± 0.0000 & 0.5315 ± 0.0159 & 0.5481 ± 0.0026 & 0.5796 ± 0.0026 & 0.4907 ± 0.0262 & 0.5000 ± 0.0000 \\
 & F1 & \cellcolor{secondblue}0.5850 ± 0.0244 & 0.5446 ± 0.0045 & \cellcolor{bestblue}0.5972 ± 0.0028 & 0.5184 ± 0.0000 & 0.4656 ± 0.0248 & 0.4546 ± 0.0483 & 0.5533 ± 0.0034 & 0.4837 ± 0.0266 & 0.4327 ± 0.0000 \\
 & NMI & \cellcolor{bestblue}0.0403 ± 0.0221 & 0.0159 ± 0.0073 & \cellcolor{secondblue}0.0275 ± 0.0016 & 0.0015 ± 0.0000 & 0.0084 ± 0.0060 & 0.0095 ± 0.0011 & 0.0266 ± 0.0015 & 0.0024 ± 0.0000 & 0.0000 ± 0.0000 \\
 & ARI & \cellcolor{bestblue}0.0380 ± 0.0197 & 0.0117 ± 0.0063 & \cellcolor{secondblue}0.0352 ± 0.0022 & -0.0035 ± 0.0000 & 0.0021 ± 0.0037 & 0.0050 ± 0.0010 & 0.0212 ± 0.0017 & -0.0022 ± 0.0000 & -0.0029 ± 0.0000 \\
\midrule

\multirow{4}{*}{StandWalkJump} & ACC & \cellcolor{bestblue}0.7556 ± 0.0831 & 0.5556 ± 0.0629 & 0.5333 ± 0.0000 & 0.4667 ± 0.0000 & 0.5111 ± 0.1133 & 0.4000 ± 0.0000 & 0.4667 ± 0.0000 & 0.1333 ± 0.0000 & \cellcolor{secondblue}0.6667 ± 0.0000 \\
 & F1 & \cellcolor{bestblue}0.7382 ± 0.1048 & \cellcolor{secondblue}0.5729 ± 0.0049 & 0.5238 ± 0.0000 & 0.5422 ± 0.0616 & 0.4582 ± 0.1094 & 0.3119 ± 0.0000 & 0.4550 ± 0.0000 & 0.0784 ± 0.0000 & 0.5342 ± 0.0000 \\
 & NMI & \cellcolor{secondblue}0.5189 ± 0.1349 & 0.3630 ± 0.0216 & 0.2928 ± 0.0000 & 0.2197 ± 0.0144 & 0.2250 ± 0.1371 & 0.0618 ± 0.0000 & 0.2392 ± 0.0000 & 0.3198 ± 0.0000 & \cellcolor{bestblue}0.5215 ± 0.0000 \\
 & ARI & \cellcolor{secondblue}0.3709 ± 0.1810 & 0.1158 ± 0.0130 & 0.1781 ± 0.0000 & 0.0734 ± 0.0018 & 0.0765 ± 0.1536 & -0.0784 ± 0.0000 & 0.0708 ± 0.0000 & 0.0973 ± 0.0000 & \cellcolor{bestblue}0.3922 ± 0.0000 \\
\bottomrule
\end{tabular}
}
\caption{Clustering performance of different methods on 15 MTS datasets evaluated by ACC (↑), F1 (↑), NMI (↑) and ARI (↑) metrics. The \colorbox{bestblue}{\textbf{Best}} and \colorbox{secondblue}{\textbf{second-best}} results are highlighted by cell background.}
\label{tab:comparison_all_metrics}
\end{table*}

\subsection{Clustering Performance Evaluation}
As Table~\ref{tab:comparison_all_metrics} shows, EMTC demonstrates superior clustering performance compared to eight baselines across 15 benchmark datasets. Achieving the highest average rank, EMTC consistently outperforms all baselines on most datasets and evaluation metrics, including ACC, F1, NMI, and ARI. Notably, it exhibits substantial improvements over the strongest competitors, i.e., FCACC and FEI. The advantages are particularly evident across diverse dataset characteristics: on long-sequence datasets like StandWalkJump, EMTC achieves significant performance gains; for datasets with balanced class distributions such as MotorImagery, it delivers notable enhancements; and on complex multi-class datasets such as Cricket (12 clusters), EMTC maintains robust performance in both clustering accuracy and mutual information metrics. This consistent superiority across all four evaluation dimensions confirms EMTC's effectiveness in handling varied temporal patterns and cluster structures.

\begin{figure}[ht]
\centering
\includegraphics[width=0.48\textwidth]{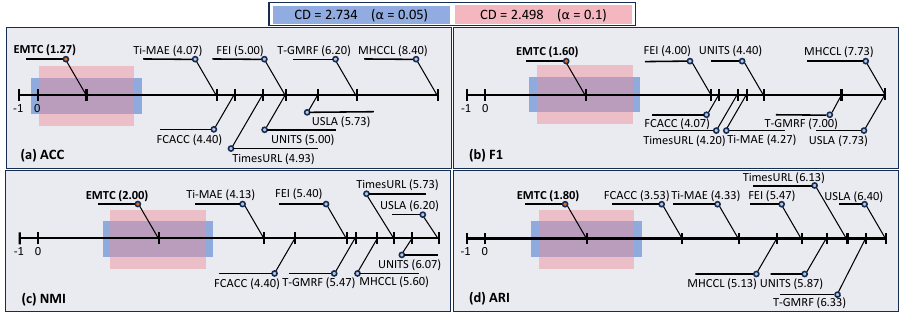}
\caption{BD test (a), (b), (c) and (d) based on the average ranks in ACC, F1, NMI and ARI presented in Table~\ref{tab:comparison_all_metrics}.}\label{significance}
\end{figure}

\subsection{Significance Test}
To statistically validate the superiority of the proposed EMTC framework, we conducted Bonferroni-Dunn tests on the clustering results from Table~\ref{tab:comparison_all_metrics}. The critical difference diagrams in Figure~\ref{significance} confirm that EMTC consistently achieves the best average ranks across all evaluation metrics, with values of 1.27 for ACC, 1.60 for F1, 2.00 for NMI, and 1.80 for ARI. Under critical difference values of 2.734 ($\alpha = 0.05$) and 2.498 ($\alpha = 0.1$) evaluated over eight baselines and 15 datasets, EMTC significantly outperforms most competing methods, as their average ranks fall outside the CD intervals. These results provide rigorous statistical evidence that EMTC establishes a new state-of-the-art in MTS clustering.

\begin{table}[!t]
\centering

\resizebox{0.47\textwidth}{!}{
\begin{tabular}{l|cc|cccc}
\toprule
\multirow{2}{*}{Datasets} & \multicolumn{2}{c|}{Components} & \multicolumn{4}{c}{Metrics} \\
\cmidrule(lr){2-3} \cmidrule(lr){4-7}
                         & IVM & MEV & ACC & F1 & NMI & ARI \\  
\midrule
\multirow{4}{*}{BasicMotions} & $\checkmark$ & $\checkmark$ & 0.9083 ± 0.0656\textcolor{white}{\ddag} & 0.9057 ± 0.0656\textcolor{white}{\ddag} & 0.8391 ± 0.1189\textcolor{white}{\ddag}
& 0.7910 ± 0.1490\textcolor{white}{\ddag}\\
 &  & $\checkmark$ & 0.7000 ± 0.1080\textcolor{blue}{\ddag} & 0.7831 ± 0.0549\textcolor{blue}{\ddag} & 0.6187 ± 0.0848\textcolor{blue}{\ddag} & 0.4799 ± 0.1214\textcolor{blue}{\ddag}\\
 & $\checkmark$ &  & 0.8750 ± 0.0354\textcolor{blue}{\ddag} & 0.8749 ± 0.0363\textcolor{blue}{\ddag} & 0.7778 ± 0.0377\textcolor{blue}{\ddag} & 0.7142 ± 0.0655\textcolor{blue}{\ddag} \\
 &  &  & 0.3167 ± 0.0236\textcolor{blue}{\ddag} & 0.2958 ± 0.0464\textcolor{blue}{\ddag} & 0.1748 ± 0.0281\textcolor{blue}{\ddag} & 0.0115 ± 0.0146\textcolor{blue}{\ddag} \\ 
\midrule

\multirow{4}{*}{Cricket} & $\checkmark$ & $\checkmark$ & 0.5972 ± 0.0227\textcolor{white}{\ddag} & 0.6317 ± 0.0366\textcolor{white}{\ddag} & 0.6574 ± 0.0273\textcolor{white}{\ddag} & 0.3436 ± 0.0303\textcolor{white}{\ddag} \\
 &  & $\checkmark$ & 0.5139 ± 0.0113\textcolor{blue}{\ddag} & 0.5423 ± 0.0524\textcolor{blue}{\ddag} & 0.6121 ± 0.0141\textcolor{blue}{\ddag} & 0.2747 ± 0.0097\textcolor{blue}{\ddag} \\
 & $\checkmark$ &  & 0.5556 ± 0.0300\textcolor{blue}{\ddag} & 0.5664 ± 0.0179\textcolor{blue}{\ddag} & 0.6433 ± 0.0347\textcolor{blue}{\ddag} & 0.3352 ± 0.0535\textcolor{blue}{\ddag} \\
 &  &  & 0.5417 ± 0.1195\textcolor{blue}{\ddag} & 0.5895 ± 0.1296\textcolor{blue}{\ddag} & 0.6443 ± 0.0809\textcolor{blue}{\ddag} & 0.3219 ± 0.1186\textcolor{blue}{\ddag} \\ 
\midrule

\multirow{4}{*}{DuckDuckGeese} & $\checkmark$ & $\checkmark$ & 0.4800 ± 0.0163\textcolor{white}{\ddag} & 0.4917 ± 0.0357\textcolor{white}{\ddag} & 0.2719 ± 0.0340\textcolor{white}{\ddag} & 0.1239 ± 0.0350\textcolor{white}{\ddag} \\
 &  & $\checkmark$ & 0.4400 ± 0.0163\textcolor{blue}{\ddag} & 0.4259 ± 0.0241\textcolor{blue}{\ddag} & 0.2293 ± 0.0074\textcolor{blue}{\ddag} & 0.0924 ± 0.0110\textcolor{blue}{\ddag} \\
 & $\checkmark$ &  & 0.4333 ± 0.0094\textcolor{blue}{\ddag} & 0.4449 ± 0.0238\textcolor{blue}{\ddag} & 0.2153 ± 0.0195\textcolor{blue}{\ddag} & 0.0836 ± 0.0125\textcolor{blue}{\ddag} \\
 &  &  & 0.2667 ± 0.0094\textcolor{blue}{\ddag} & 0.2227 ± 0.0041\textcolor{blue}{\ddag} & 0.1655 ± 0.0013\textcolor{blue}{\ddag} & -0.0010 ± 0.0011\textcolor{blue}{\ddag} \\ 
\midrule

\multirow{4}{*}{EigenWorms} & $\checkmark$ & $\checkmark$ & 0.4707 ± 0.0095\textcolor{white}{\ddag} & 0.3713 ± 0.0549\textcolor{white}{\ddag} & 0.1538 ± 0.0275\textcolor{white}{\ddag} & 0.1012 ± 0.0435\textcolor{white}{\ddag} \\
 &  & $\checkmark$ & 0.4233 ± 0.0015\textcolor{blue}{\ddag} & 0.3223 ± 0.0311\textcolor{blue}{\ddag} & 0.1566 ± 0.0135\textcolor{white}{\ddag} & 0.1011 ± 0.0335\textcolor{blue}{\ddag} \\
 & $\checkmark$ &  & 0.4245 ± 0.0011\textcolor{blue}{\ddag} & 0.3112 ± 0.0034\textcolor{blue}{\ddag} & 0.1223 ± 0.0325\textcolor{blue}{\ddag} & 0.1011 ± 0.0000\textcolor{blue}{\ddag} \\
 &  &  & 0.2235 ± 0.0032\textcolor{blue}{\ddag} & 0.1812 ± 0.0014\textcolor{blue}{\ddag} & 0.1122 ± 0.0125\textcolor{blue}{\ddag} & 0.0073 ± 0.0012\textcolor{blue}{\ddag} \\ 
\midrule

\multirow{4}{*}{Epilepsy} & $\checkmark$ & $\checkmark$ & 0.5556 ± 0.0239\textcolor{white}{\ddag} & 0.5519 ± 0.0276\textcolor{white}{\ddag} & 0.2783 ± 0.0182\textcolor{white}{\ddag} & 0.2068 ± 0.0273\textcolor{white}{\ddag} \\
 &  & $\checkmark$ & 0.5024 ± 0.0292\textcolor{blue}{\ddag} & 0.4595 ± 0.0601\textcolor{blue}{\ddag} & 0.2366 ± 0.0526\textcolor{blue}{\ddag} & 0.1853 ± 0.0365\textcolor{blue}{\ddag} \\
 & $\checkmark$ &  & 0.4928 ± 0.0059\textcolor{blue}{\ddag} & 0.4850 ± 0.0093\textcolor{blue}{\ddag} & 0.1981 ± 0.0414\textcolor{blue}{\ddag} & 0.1257 ± 0.0134\textcolor{blue}{\ddag} \\
 &  &  & 0.3961 ± 0.0326\textcolor{blue}{\ddag} & 0.3437 ± 0.0625\textcolor{blue}{\ddag} & 0.0767 ± 0.0184\textcolor{blue}{\ddag} & 0.0443 ± 0.0142\textcolor{blue}{\ddag} \\ 
\midrule

\multirow{4}{*}{\shortstack[l]{Finger\\Movements}} & $\checkmark$ & $\checkmark$ & 0.5967 ± 0.0094\textcolor{white}{\ddag} & 0.5892 ± 0.0151\textcolor{white}{\ddag} & 0.0320 ± 0.0030\textcolor{white}{\ddag} & 0.0287 ± 0.0071\textcolor{white}{\ddag} \\
 &  & $\checkmark$ & 0.6067 ± 0.0047\textcolor{white}{\ddag} & 0.5982 ± 0.0131\textcolor{white}{\ddag} & 0.0404 ± 0.0031\textcolor{white}{\ddag} & 0.0367 ± 0.0032\textcolor{white}{\ddag} \\
 & $\checkmark$ &  & 0.6033 ± 0.0125\textcolor{white}{\ddag} & 0.5933 ± 0.0199\textcolor{white}{\ddag} & 0.0402 ± 0.0120\textcolor{white}{\ddag} & 0.0345 ± 0.0105\textcolor{white}{\ddag} \\
 &  &  & 0.5233 ± 0.0262\textcolor{blue}{\ddag} & 0.4213 ± 0.0970\textcolor{blue}{\ddag} & 0.0149 ± 0.0144\textcolor{blue}{\ddag} & 0.0010 ± 0.0025\textcolor{blue}{\ddag} \\ 
\midrule

\multirow{4}{*}{\shortstack[l]{HandMovement\\ Direction}} & $\checkmark$ & $\checkmark$ & 0.4640 ± 0.0064\textcolor{white}{\ddag} & 0.3737 ± 0.0262\textcolor{white}{\ddag} & 0.0914 ± 0.0137\textcolor{white}{\ddag} & 0.0767 ± 0.0068\textcolor{white}{\ddag} \\
 &  & $\checkmark$ & 0.4730 ± 0.0191\textcolor{white}{\ddag} & 0.4158 ± 0.0291\textcolor{white}{\ddag} & 0.0982 ± 0.0113\textcolor{white}{\ddag} & 0.0703 ± 0.0374\textcolor{blue}{\ddag} \\
 & $\checkmark$ &  & 0.4414 ± 0.0127\textcolor{blue}{\ddag} & 0.4147 ± 0.0562\textcolor{white}{\ddag} & 0.1021 ± 0.0246\textcolor{white}{\ddag} & 0.0682 ± 0.0204\textcolor{blue}{\ddag} \\
 &  &  & 0.3784 ± 0.0398\textcolor{blue}{\ddag} & 0.3398 ± 0.0288\textcolor{blue}{\ddag} & 0.0586 ± 0.0260\textcolor{blue}{\ddag} & 0.0118 ± 0.0239\textcolor{blue}{\ddag} \\ 
\midrule

\multirow{4}{*}{Heartbeat} & $\checkmark$ & $\checkmark$ & 0.7463 ± 0.0040\textcolor{white}{\ddag} & 0.6134 ± 0.0329\textcolor{white}{\ddag} & 0.0733 ± 0.0123\textcolor{white}{\ddag} & 0.1531 ± 0.0273\textcolor{white}{\ddag} \\
 &  & $\checkmark$ & 0.7415 ± 0.0040\textcolor{blue}{\ddag} & 0.5621 ± 0.0637\textcolor{blue}{\ddag} & 0.0764 ± 0.0104\textcolor{white}{\ddag} & 0.1127 ± 0.0434\textcolor{blue}{\ddag} \\
 & $\checkmark$ &  & 0.7447 ± 0.0122\textcolor{blue}{\ddag} & 0.6336 ± 0.0133\textcolor{white}{\ddag} & 0.0810 ± 0.0106\textcolor{white}{\ddag} & 0.1675 ± 0.0020\textcolor{white}{\ddag} \\
 &  &  & 0.7138 ± 0.0023\textcolor{blue}{\ddag} & 0.8330 ± 0.0016\textcolor{white}{\ddag} & 0.0158 ± 0.0028\textcolor{blue}{\ddag} & -0.0098 ± 0.0027\textcolor{blue}{\ddag} \\ 
\midrule

\multirow{4}{*}{MotorImagery} & $\checkmark$ & $\checkmark$ & 0.6500 ± 0.0216\textcolor{white}{\ddag} & 0.6452 ± 0.0213\textcolor{white}{\ddag} & 0.0737 ± 0.0229\textcolor{white}{\ddag} & 0.0831 ± 0.0274\textcolor{white}{\ddag} \\
 &  & $\checkmark$ & 0.4400 ± 0.0163\textcolor{blue}{\ddag} & 0.4259 ± 0.0241\textcolor{blue}{\ddag} & 0.2293 ± 0.0074\textcolor{white}{\ddag} & 0.0924 ± 0.0110\textcolor{white}{\ddag} \\
 & $\checkmark$ &  & 0.4333 ± 0.0094\textcolor{blue}{\ddag} & 0.4449 ± 0.0238\textcolor{blue}{\ddag} & 0.2153 ± 0.0195\textcolor{white}{\ddag} & 0.0836 ± 0.0125\textcolor{white}{\ddag} \\
 &  &  & 0.5533 ± 0.0309\textcolor{blue}{\ddag} & 0.4972 ± 0.1005\textcolor{blue}{\ddag} & 0.0237 ± 0.0085\textcolor{blue}{\ddag} & 0.0088 ± 0.0067\textcolor{blue}{\ddag} \\ 
\midrule

\multirow{4}{*}{NATOPS} & $\checkmark$ & $\checkmark$ & 0.6185 ± 0.0094\textcolor{white}{\ddag} & 0.6205 ± 0.0130\textcolor{white}{\ddag} & 0.5503 ± 0.0058\textcolor{white}{\ddag} & 0.3966 ± 0.0155\textcolor{white}{\ddag} \\
 &  & $\checkmark$ & 0.5093 ± 0.0664\textcolor{blue}{\ddag} & 0.4951 ± 0.0726\textcolor{blue}{\ddag} & 0.4208 ± 0.0544\textcolor{blue}{\ddag} & 0.2765 ± 0.0590\textcolor{blue}{\ddag} \\
 & $\checkmark$ &  & 0.5296 ± 0.0502\textcolor{blue}{\ddag} & 0.5605 ± 0.0912\textcolor{blue}{\ddag} & 0.4541 ± 0.0755\textcolor{blue}{\ddag} & 0.2975 ± 0.0803\textcolor{blue}{\ddag} \\
 &  &  & 0.4204 ± 0.1089\textcolor{blue}{\ddag} & 0.4145 ± 0.1189\textcolor{blue}{\ddag} & 0.3708 ± 0.1074\textcolor{blue}{\ddag} & 0.2077 ± 0.1024\textcolor{blue}{\ddag} \\ 
\midrule

\multirow{4}{*}{PEMS-SF} & $\checkmark$ & $\checkmark$ & 0.5780 ± 0.0164\textcolor{white}{\ddag} & 0.6215 ± 0.0398\textcolor{white}{\ddag} & 0.5316 ± 0.0347\textcolor{white}{\ddag} & 0.3611 ± 0.0277\textcolor{white}{\ddag} \\
 &  & $\checkmark$ & 0.4566 ± 0.0000\textcolor{blue}{\ddag} & 0.4149 ± 0.0012\textcolor{blue}{\ddag} & 0.3646 ± 0.0000\textcolor{blue}{\ddag} & 0.2027 ± 0.0000\textcolor{blue}{\ddag} \\
 & $\checkmark$ &  & 0.4509 ± 0.0324\textcolor{blue}{\ddag} & 0.5077 ± 0.0133\textcolor{blue}{\ddag} & 0.4465 ± 0.0000\textcolor{blue}{\ddag} & 0.2401 ± 0.0000\textcolor{blue}{\ddag} \\
 &  &  & 0.2543 ± 0.0544\textcolor{blue}{\ddag} & 0.2037 ± 0.0695\textcolor{blue}{\ddag} & 0.1279 ± 0.0587\textcolor{blue}{\ddag} & 0.0404 ± 0.0525\textcolor{blue}{\ddag} \\ 
\midrule

\multirow{4}{*}{RacketSports} & $\checkmark$ & $\checkmark$ & 0.4715 ± 0.0135\textcolor{white}{\ddag} & 0.4657 ± 0.0695\textcolor{white}{\ddag} & 0.2004 ± 0.0520\textcolor{white}{\ddag} & 0.1235 ± 0.0372\textcolor{white}{\ddag} \\
 &  & $\checkmark$ & 0.4737 ± 0.0387\textcolor{white}{\ddag} & 0.4778 ± 0.0443\textcolor{white}{\ddag} & 0.1897 ± 0.0601\textcolor{blue}{\ddag} & 0.1209 ± 0.0464\textcolor{blue}{\ddag} \\
 & $\checkmark$ &  & 0.4868 ± 0.0327\textcolor{white}{\ddag} & 0.4899 ± 0.0560\textcolor{white}{\ddag} & 0.2044 ± 0.0425\textcolor{white}{\ddag} & 0.1517 ± 0.0214\textcolor{white}{\ddag} \\
 &  &  & 0.3399 ± 0.0487\textcolor{blue}{\ddag} & 0.2931 ± 0.0673\textcolor{blue}{\ddag} & 0.0757 ± 0.0373\textcolor{blue}{\ddag} & 0.0290 ± 0.0393\textcolor{blue}{\ddag} \\ 
\midrule

\multirow{4}{*}{\shortstack[l]{SelfRegulation\\SCP1}} & $\checkmark$ & $\checkmark$ & 0.8510 ± 0.0126\textcolor{white}{\ddag} & 0.8500 ± 0.0135\textcolor{white}{\ddag} & 0.4146 ± 0.0207\textcolor{white}{\ddag} & 0.4916 ± 0.0355\textcolor{white}{\ddag} \\
 &  & $\checkmark$ & 0.7782 ± 0.0121\textcolor{blue}{\ddag} & 0.7757 ± 0.0105\textcolor{blue}{\ddag} & 0.2619 ± 0.0394\textcolor{blue}{\ddag} & 0.3078 ± 0.0268\textcolor{blue}{\ddag} \\
 & $\checkmark$ &  & 0.7713 ± 0.0255\textcolor{blue}{\ddag} & 0.7706 ± 0.0254\textcolor{blue}{\ddag} & 0.2322 ± 0.0498\textcolor{blue}{\ddag} & 0.2947 ± 0.0567\textcolor{blue}{\ddag} \\
 &  &  & 0.7531 ± 0.1754\textcolor{blue}{\ddag} & 0.7006 ± 0.2497\textcolor{blue}{\ddag} & 0.3178 ± 0.2048\textcolor{blue}{\ddag} & 0.3783 ± 0.2675\textcolor{blue}{\ddag} \\ 
\midrule

\multirow{4}{*}{\shortstack[l]{SelfRegulation\\SCP2}} & $\checkmark$ & $\checkmark$ & 0.6000 ± 0.0253\textcolor{white}{\ddag} & 0.5850 ± 0.0244\textcolor{white}{\ddag} & 0.0403 ± 0.0221\textcolor{white}{\ddag} & 0.0380 ± 0.0197\textcolor{white}{\ddag} \\
 &  & $\checkmark$ & 0.5963 ± 0.0105\textcolor{blue}{\ddag} & 0.5953 ± 0.0095\textcolor{white}{\ddag} & 0.0277 ± 0.0067\textcolor{blue}{\ddag} & 0.0322 ± 0.0085\textcolor{blue}{\ddag} \\
 & $\checkmark$ &  & 0.5815 ± 0.0069\textcolor{blue}{\ddag} & 0.5658 ± 0.0143\textcolor{blue}{\ddag} & 0.0238 ± 0.0022\textcolor{blue}{\ddag} & 0.0221 ± 0.0041\textcolor{blue}{\ddag} \\
 &  &  & 0.5241 ± 0.0233\textcolor{blue}{\ddag} & 0.4710 ± 0.0921\textcolor{blue}{\ddag} & 0.0033 ± 0.0040\textcolor{blue}{\ddag} & 0.0007 ± 0.0046\textcolor{blue}{\ddag}
 \\ 
\midrule

\multirow{4}{*}{StandWalkJump} & $\checkmark$ & $\checkmark$ & 0.7556 ± 0.0831\textcolor{white}{\ddag} & 0.7382 ± 0.1048\textcolor{white}{\ddag} & 0.5189 ± 0.1349\textcolor{white}{\ddag} & 0.3709 ± 0.1810\textcolor{white}{\ddag} \\
 &  & $\checkmark$ & 0.6000 ± 0.0544\textcolor{blue}{\ddag} & 0.5681 ± 0.0646\textcolor{blue}{\ddag} & 0.2826 ± 0.0451\textcolor{blue}{\ddag} & 0.1274 ± 0.0598\textcolor{blue}{\ddag} \\
 & $\checkmark$ &  & 0.6444 ± 0.0629\textcolor{blue}{\ddag} & 0.6165 ± 0.0688\textcolor{blue}{\ddag} & 0.3739 ± 0.0859\textcolor{blue}{\ddag} & 0.1832 ± 0.1350\textcolor{blue}{\ddag} \\
 &  &  & 0.4222 ± 0.0314\textcolor{blue}{\ddag} & 0.4321 ± 0.0175\textcolor{blue}{\ddag} & 0.2229 ± 0.0109\textcolor{blue}{\ddag} &  0.0120 ± 0.0149\textcolor{blue}{\ddag}\\ 
\bottomrule
\end{tabular}
}
\caption{Ablation study of key components. The symbol ``$\checkmark$'' denotes included components, and the symbol ``\textcolor{blue}{\ddag}'' indicates performance degradation compared to the complete EMTC.}
\label{tb:ablation}
\end{table}

\begin{figure}[t]
\centering
\includegraphics[width=\columnwidth]{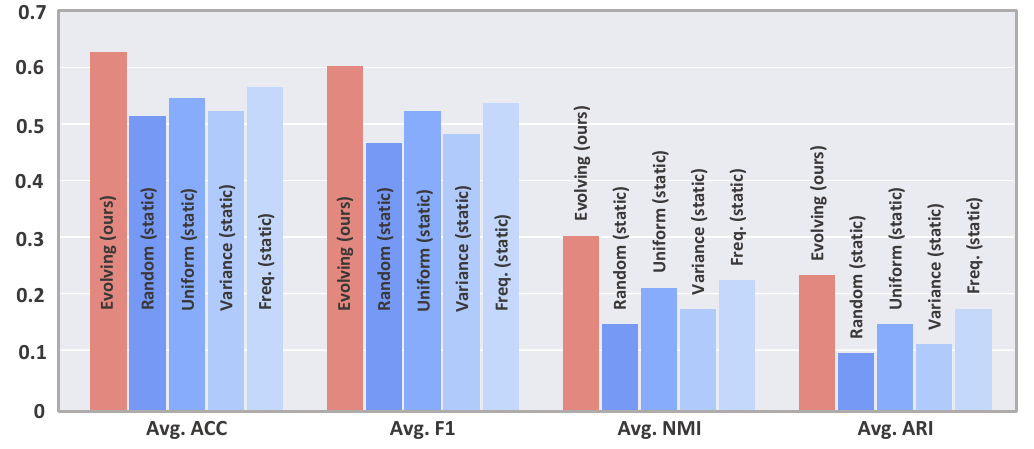}
\caption{Clustering performance of EMTC+Evolving Masking vs. EMTC+Static Masking.}
\label{mask_compare}
\end{figure}

\subsection{Masking Strategies Comparison}
Efficacy of the proposed evolving masking is validated by comparing it with four conventional static masking variants, including Random, Uniform, Variance-based, and Frequency-based masking, through replacement of the IVM module of EMTC. As shown in Figure~\ref{mask_compare}, evolving masking consistently outperforms all static counterparts across all evaluation metrics, with average performance computed over all 15 benchmark datasets. Although Variance-based and Frequency-based methods incorporate data statistics, their static nature limits their adaptability to different clustering tasks. In contrast, our approach progressively refines its focus on cluster-salient timestamps through content-aware importance evaluation described bu Eq.~(\ref{importance_evaluation}), enabling precise redundancy suppression while preserving discriminative and clustering-friendly patterns.

\subsection{Convergence Analysis}
\begin{figure}[t]
\centering
\includegraphics[width=\columnwidth]{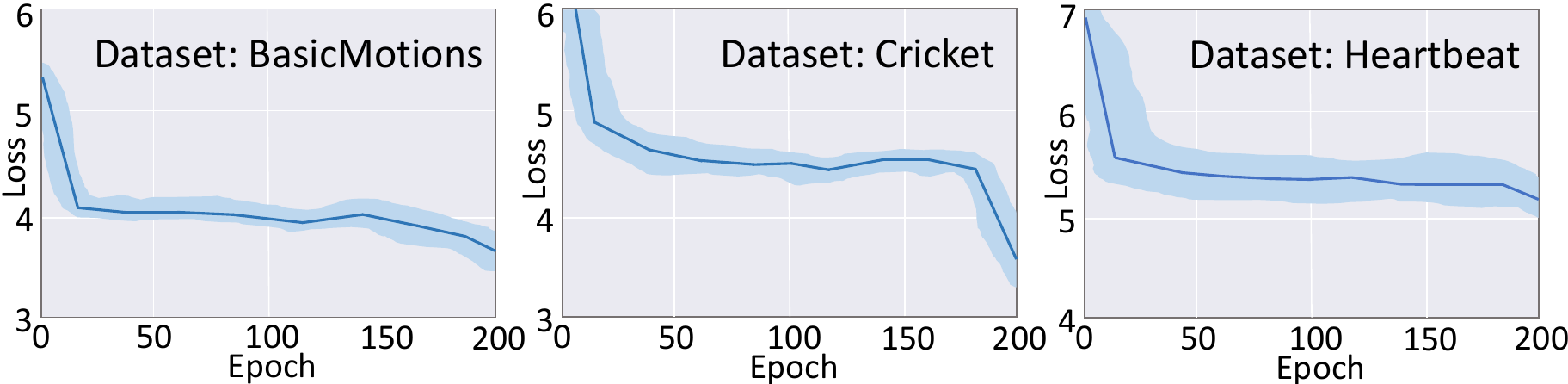}
\caption{Loss function values during EMTC training.}
\label{loss}
\end{figure}

Figure~\ref{loss} illustrates the convergence behavior across three representative datasets, showing smooth and stable optimization with the total loss consistently decreasing throughout training. The convergence characteristics vary appropriately with dataset properties: BasicMotions with balanced classes exhibits faster initial convergence, Cricket with more classes shows gradual stabilization, and Heartbeat with longer sequences maintains steady decline. This stable convergence validates the effectiveness of our joint optimization strategy Eq.~(\ref{eq:13}), where the synergy between CRL and CMC ensures progressive refinement of both representation quality and cluster structures, confirming the framework's robustness in learning representations across diverse MTS clustering scenarios.

\subsection{Ablation Study of Key Components}
The results in Table~\ref{tb:ablation} demonstrates that both IVM and MEV are essential for EMTC's superior performance. The complete model consistently outperforms all ablated variants, with the absence of either component causing significant performance degradation. IVM generally contributes more substantially than MEV, particularly on long-sequence datasets like \textit{StandWalkJump} where temporal redundancy is prominent. It is worth noting that the performance gap between the full model and the version without both MEV and IVM became more significant, confirming the synergistic effect of the two key components.
Overall, the ablation validates the coupling design: IVM dynamically suppresses redundant timestamps while MEV comprehensively captures multi-view patterns to consolidate masking effectiveness and enhance cluster separation.

\begin{table}[!t]
\centering
\resizebox{0.47\textwidth}{!}{
\begin{tabular}{l|ccc|cccc}
\toprule
\multirow{2}{*}{Datasets} & \multicolumn{3}{c|}{Loss Terms} & \multicolumn{4}{c}{Metrics} \\
\cmidrule(lr){2-4} \cmidrule(lr){5-8}
                         & Intra & Inter & Contra & ACC & F1 & NMI & ARI \\  
\midrule
\multirow{5}{*}{BasicMotions} & $\checkmark$ & $\checkmark$ & $\checkmark$ & 0.9083 ± 0.0656\textcolor{white}{\ddag} & 0.9057 ± 0.0656\textcolor{white}{\ddag} & 0.8391 ± 0.1189\textcolor{white}{\ddag}
& 0.7910 ± 0.1490\textcolor{white}{\ddag}\\
 &  & $\checkmark$ & $\checkmark$ & 0.6000 ± 0.0000\textcolor{blue}{\ddag} & 0.6178 ± 0.0626\textcolor{blue}{\ddag} & 0.4772 ± 0.0886\textcolor{blue}{\ddag} & 0.3317 ± 0.0618\textcolor{blue}{\ddag}\\
 & $\checkmark$ & & $\checkmark$ & 0.5833 ± 0.1027\textcolor{blue}{\ddag} & 0.5787 ± 0.0902\textcolor{blue}{\ddag} & 0.3817 ± 0.1417\textcolor{blue}{\ddag} & 0.2311 ± 0.1073\textcolor{blue}{\ddag} \\
 &  & & $\checkmark$ &  0.5917 ± 0.1124\textcolor{blue}{\ddag} & 0.6377 ± 0.1069\textcolor{blue}{\ddag} & 0.4409 ± 0.1413\textcolor{blue}{\ddag} & 0.3073 ± 0.1432\textcolor{blue}{\ddag} \\ 
  & $\checkmark$ & $\checkmark$ &  & 0.5750 ± 0.0540\textcolor{blue}{\ddag} & 0.5662 ± 0.0538\textcolor{blue}{\ddag} & 0.2943 ± 0.0912\textcolor{blue}{\ddag} & 0.1908 ± 0.0969\textcolor{blue}{\ddag}\\
\midrule

\multirow{5}{*}{Cricket} & $\checkmark$ & $\checkmark$ & $\checkmark$ & 0.5972 ± 0.0227\textcolor{white}{\ddag} & 0.6317 ± 0.0366\textcolor{white}{\ddag} & 0.6574 ± 0.0273\textcolor{white}{\ddag}
& 0.3436 ± 0.0303\textcolor{white}{\ddag}\\
&  & $\checkmark$ & $\checkmark$ & 0.4722 ± 0.0113\textcolor{blue}{\ddag} & 0.5098 ± 0.0377\textcolor{blue}{\ddag} & 0.5417 ± 0.0199\textcolor{blue}{\ddag} & 0.2071 ± 0.0128\textcolor{blue}{\ddag}\\
& $\checkmark$ & & $\checkmark$ & 0.4491 ± 0.0346\textcolor{blue}{\ddag} & 0.4920 ± 0.0354\textcolor{blue}{\ddag} & 0.5490 ± 0.0454\textcolor{blue}{\ddag} & 0.2108 ± 0.0563\textcolor{blue}{\ddag} \\
&  & & $\checkmark$ &  0.4861 ± 0.0600\textcolor{blue}{\ddag} & 0.5408 ± 0.0112\textcolor{blue}{\ddag} & 0.5751 ± 0.0491\textcolor{blue}{\ddag} & 0.2448 ± 0.0692\textcolor{blue}{\ddag} \\ 
& $\checkmark$ & $\checkmark$ &  & 0.4676 ± 0.0559\textcolor{blue}{\ddag} & 0.4629 ± 0.0380\textcolor{blue}{\ddag} & 0.5423 ± 0.0650\textcolor{blue}{\ddag} & 0.1956 ± 0.0682\textcolor{blue}{\ddag}\\
\midrule

\multirow{5}{*}{DuckDuckGeese} & $\checkmark$ & $\checkmark$ & $\checkmark$ & 0.4800 ± 0.0163\textcolor{white}{\ddag} & 0.4917 ± 0.0357\textcolor{white}{\ddag} & 0.2719 ± 0.0340\textcolor{white}{\ddag}
& 0.1239 ± 0.0350\textcolor{white}{\ddag}\\
&  & $\checkmark$ & $\checkmark$ & 0.4533 ± 0.0411\textcolor{blue}{\ddag} & 0.4502 ± 0.0303\textcolor{blue}{\ddag} & 0.2479 ± 0.0355\textcolor{blue}{\ddag} & 0.0998 ± 0.0435\textcolor{blue}{\ddag}\\
& $\checkmark$ & & $\checkmark$ & 0.4333 ± 0.0340\textcolor{blue}{\ddag} & 0.4382 ± 0.0293\textcolor{blue}{\ddag} & 0.2185 ± 0.0620\textcolor{blue}{\ddag} & 0.0852 ± 0.0467\textcolor{blue}{\ddag} \\
&  & & $\checkmark$ & 0.4333 ± 0.0340\textcolor{blue}{\ddag} & 0.4382 ± 0.0293\textcolor{blue}{\ddag} & 0.2185 ± 0.0620\textcolor{blue}{\ddag} & 0.0852 ± 0.0467\textcolor{blue}{\ddag} \\
& $\checkmark$ & $\checkmark$ &  & 0.4000 ± 0.0163\textcolor{blue}{\ddag} & 0.3747 ± 0.0209\textcolor{blue}{\ddag} & 0.2013 ± 0.0378\textcolor{blue}{\ddag} & 0.0616 ± 0.0220\textcolor{blue}{\ddag}\\
\midrule

\multirow{5}{*}{EigenWorms} & $\checkmark$ & $\checkmark$ & $\checkmark$ & 0.4707 ± 0.0095\textcolor{white}{\ddag} & 0.3713 ± 0.0549\textcolor{white}{\ddag} & 0.1538 ± 0.0275\textcolor{white}{\ddag}
& 0.1012 ± 0.0435\textcolor{white}{\ddag}\\
&  & $\checkmark$ & $\checkmark$ & 0.4529 ± 0.0406\textcolor{blue}{\ddag} & 0.2947 ± 0.0196\textcolor{blue}{\ddag} & 0.1260 ± 0.0360\textcolor{blue}{\ddag} & 0.0880 ± 0.0735\textcolor{blue}{\ddag}\\
& $\checkmark$ & & $\checkmark$ & 0.4707 ± 0.0320\textcolor{blue}{\ddag} & 0.3671 ± 0.0648\textcolor{blue}{\ddag} & 0.1683 ± 0.0688\textcolor{white}{\ddag} & 0.1257 ± 0.0843\textcolor{white}{\ddag} \\
&  & & $\checkmark$ & 0.4860 ± 0.0200\textcolor{white}{\ddag} & 0.3995 ± 0.0867\textcolor{white}{\ddag} & 0.1803 ± 0.0431\textcolor{white}{\ddag} & 0.1245 ± 0.0770\textcolor{white}{\ddag} \\
& $\checkmark$ & $\checkmark$ &  & 0.4936 ± 0.0036\textcolor{white}{\ddag} & 0.4058 ± 0.0855\textcolor{white}{\ddag} & 0.2010 ± 0.0517\textcolor{white}{\ddag} & 0.1677 ± 0.0220\textcolor{white}{\ddag} \\
\midrule

\multirow{5}{*}{Epilepsy} & $\checkmark$ & $\checkmark$ & $\checkmark$ & 0.5556 ± 0.0239\textcolor{white}{\ddag} & 0.5519 ± 0.0276\textcolor{white}{\ddag} & 0.2783 ± 0.0182\textcolor{white}{\ddag}
& 0.2068 ± 0.0273\textcolor{white}{\ddag}\\
&  & $\checkmark$ & $\checkmark$ & 0.4855 ± 0.0313\textcolor{blue}{\ddag} & 0.4764 ± 0.0287\textcolor{blue}{\ddag} & 0.2243 ± 0.0289\textcolor{blue}{\ddag} & 0.1279 ± 0.0308\textcolor{blue}{\ddag}\\
& $\checkmark$ & & $\checkmark$ & 0.5024 ± 0.0239\textcolor{blue}{\ddag} & 0.4937 ± 0.0204\textcolor{blue}{\ddag} & 0.2200 ± 0.0383\textcolor{blue}{\ddag} & 0.1386 ± 0.0351\textcolor{blue}{\ddag} \\
&  & & $\checkmark$ & 0.5024 ± 0.0239\textcolor{blue}{\ddag} & 0.4978 ± 0.0216\textcolor{blue}{\ddag} & 0.2582 ± 0.0191\textcolor{blue}{\ddag} & 0.1645 ± 0.0452\textcolor{blue}{\ddag} \\
& $\checkmark$ & $\checkmark$ &  & 0.4638 ± 0.0157\textcolor{blue}{\ddag} & 0.4445 ± 0.0336\textcolor{blue}{\ddag} & 0.1386 ± 0.0334\textcolor{blue}{\ddag} & 0.0895 ± 0.0174\textcolor{blue}{\ddag}\\
\midrule

\multirow{5}{*}{\shortstack[l]{Finger\\Movements}} & $\checkmark$ & $\checkmark$ & $\checkmark$ & 0.5967 ± 0.0094\textcolor{white}{\ddag} & 0.5892 ± 0.0151\textcolor{white}{\ddag} & 0.0320 ± 0.0030\textcolor{white}{\ddag}
& 0.0287 ± 0.0071\textcolor{white}{\ddag}\\
&  & $\checkmark$ & $\checkmark$ & 0.6033 ± 0.0125\textcolor{white}{\ddag} & 0.5925 ± 0.0128\textcolor{white}{\ddag} & 0.0389 ± 0.0148\textcolor{white}{\ddag} & 0.0346 ± 0.0110\textcolor{white}{\ddag}\\
& $\checkmark$ & & $\checkmark$ & 0.6067 ± 0.0170\textcolor{white}{\ddag} & 0.6055 ± 0.0165\textcolor{white}{\ddag} & 0.0341 ± 0.0110\textcolor{white}{\ddag} & 0.0370 ± 0.0153\textcolor{white}{\ddag} \\
&  & & $\checkmark$ & 0.6033 ± 0.0047\textcolor{white}{\ddag} & 0.6030 ± 0.0047\textcolor{white}{\ddag} & 0.0314 ± 0.0026\textcolor{blue}{\ddag} & 0.0331 ± 0.0040\textcolor{white}{\ddag} \\
& $\checkmark$ & $\checkmark$ &  & 0.6100 ± 0.0216\textcolor{white}{\ddag} & 0.6073 ± 0.0197\textcolor{white}{\ddag} & 0.0406 ± 0.0182\textcolor{white}{\ddag} & 0.0409 ± 0.0206\textcolor{white}{\ddag} \\
\midrule

\multirow{5}{*}{\shortstack[l]{HandMovement\\ Direction}} & $\checkmark$ & $\checkmark$ & $\checkmark$ & 0.4640 ± 0.0064\textcolor{white}{\ddag} & 0.3737 ± 0.0262\textcolor{white}{\ddag} & 0.0914 ± 0.0137\textcolor{white}{\ddag}
& 0.0767 ± 0.0068\textcolor{white}{\ddag}\\
&  & $\checkmark$ & $\checkmark$ & 0.4595 ± 0.0110\textcolor{blue}{\ddag} & 0.3740 ± 0.0585\textcolor{white}{\ddag} & 0.1105 ± 0.0079\textcolor{white}{\ddag} & 0.0524 ± 0.0059\textcolor{blue}{\ddag}\\
& $\checkmark$ & & $\checkmark$ & 0.4414 ± 0.0064\textcolor{blue}{\ddag} & 0.3814 ± 0.0441\textcolor{white}{\ddag} & 0.0762 ± 0.0087\textcolor{blue}{\ddag} & 0.0447 ± 0.0115\textcolor{blue}{\ddag} \\
&  & & $\checkmark$ & 0.4730 ± 0.0382\textcolor{white}{\ddag} & 0.3959 ± 0.0570\textcolor{white}{\ddag} & 0.0984 ± 0.0358\textcolor{white}{\ddag} & 0.0776 ± 0.0335\textcolor{white}{\ddag} \\
& $\checkmark$ & $\checkmark$ &  & 0.4530 ± 0.0250\textcolor{blue}{\ddag} & 0.3427 ± 0.0378\textcolor{blue}{\ddag} & 0.1238 ± 0.0625\textcolor{white}{\ddag} & 0.0916 ± 0.0582\textcolor{white}{\ddag} \\
\midrule

\multirow{5}{*}{Heartbeat} & $\checkmark$ & $\checkmark$ & $\checkmark$ & 0.7463 ± 0.0040\textcolor{white}{\ddag} & 0.6134 ± 0.0329\textcolor{white}{\ddag} & 0.0733 ± 0.0123\textcolor{white}{\ddag} & 0.1531 ± 0.0273\textcolor{white}{\ddag}\\
&  & $\checkmark$ & $\checkmark$ & 0.7317 ± 0.0040\textcolor{blue}{\ddag} & 0.5538 ± 0.0329\textcolor{blue}{\ddag} & 0.0419 ± 0.0057\textcolor{blue}{\ddag} & 0.0947 ± 0.0221\textcolor{blue}{\ddag}\\
& $\checkmark$ & & $\checkmark$ & 0.7301 ± 0.0128\textcolor{blue}{\ddag} & 0.6503 ± 0.1340\textcolor{white}{\ddag} & 0.0422 ± 0.0186\textcolor{blue}{\ddag} & 0.0672 ± 0.0629\textcolor{blue}{\ddag} \\
&  & & $\checkmark$ & 0.7285 ± 0.0061\textcolor{blue}{\ddag} & 0.5017 ± 0.0516\textcolor{blue}{\ddag} & 0.0270 ± 0.0167\textcolor{blue}{\ddag} & 0.0571 ± 0.0399\textcolor{blue}{\ddag} \\
& $\checkmark$ & $\checkmark$ &  & 0.7333 ± 0.0046\textcolor{blue}{\ddag} & 0.5537 ± 0.0213\textcolor{blue}{\ddag} & 0.0441 ± 0.0037\textcolor{blue}{\ddag} & 0.0958 ± 0.0123\textcolor{blue}{\ddag}\\
\midrule

\multirow{5}{*}{MotorImagery} & $\checkmark$ & $\checkmark$ & $\checkmark$ & 0.6500 ± 0.0216\textcolor{white}{\ddag} & 0.6452 ± 0.0213\textcolor{white}{\ddag} & 0.0737 ± 0.0229\textcolor{white}{\ddag} & 0.0831 ± 0.0274\textcolor{white}{\ddag}\\
&  & $\checkmark$ & $\checkmark$ & 0.6067 ± 0.0309\textcolor{blue}{\ddag} & 0.5931 ± 0.0374\textcolor{blue}{\ddag} & 0.0418 ± 0.0208\textcolor{blue}{\ddag} & 0.0409 ± 0.0286\textcolor{blue}{\ddag}\\
& $\checkmark$ & & $\checkmark$ & 0.6033 ± 0.0262\textcolor{blue}{\ddag} & 0.5911 ± 0.0346\textcolor{white}{\ddag} & 0.0373 ± 0.0145\textcolor{blue}{\ddag} & 0.0369 ± 0.0229\textcolor{blue}{\ddag} \\
&  & & $\checkmark$ & 0.6567 ± 0.0386\textcolor{white}{\ddag} & 0.6498 ± 0.0407\textcolor{white}{\ddag} & 0.0853 ± 0.0408\textcolor{white}{\ddag} & 0.0957 ± 0.0521\textcolor{white}{\ddag} \\
& $\checkmark$ & $\checkmark$ &  & 0.6067 ± 0.0377\textcolor{blue}{\ddag} & 0.5890 ± 0.0453\textcolor{blue}{\ddag} & 0.0468 ± 0.0296\textcolor{blue}{\ddag} & 0.0431 ± 0.0361\textcolor{blue}{\ddag}\\
\midrule

\multirow{5}{*}{NATOPS} & $\checkmark$ & $\checkmark$ & $\checkmark$ & 0.6185 ± 0.0094\textcolor{white}{\ddag} & 0.6205 ± 0.0130\textcolor{white}{\ddag} & 0.5503 ± 0.0058\textcolor{white}{\ddag} & 0.3966 ± 0.0155\textcolor{white}{\ddag}\\
&  & $\checkmark$ & $\checkmark$ & 0.5963 ± 0.0139\textcolor{blue}{\ddag} & 0.5917 ± 0.0153\textcolor{blue}{\ddag} & 0.5071 ± 0.0463\textcolor{blue}{\ddag} & 0.3726 ± 0.0354\textcolor{blue}{\ddag}\\
& $\checkmark$ & & $\checkmark$ & 0.5093 ± 0.0328\textcolor{blue}{\ddag} & 0.5515 ± 0.0141\textcolor{blue}{\ddag} & 0.4731 ± 0.0217\textcolor{blue}{\ddag} & 0.2926 ± 0.0381\textcolor{blue}{\ddag} \\
&  & & $\checkmark$ & 0.5407 ± 0.0421\textcolor{blue}{\ddag} & 0.5583 ± 0.0223\textcolor{blue}{\ddag} & 0.4918 ± 0.0126\textcolor{blue}{\ddag} & 0.3244 ± 0.0451\textcolor{blue}{\ddag} \\
& $\checkmark$ & $\checkmark$ &  & 0.3741 ± 0.1049\textcolor{blue}{\ddag} & 0.3561 ± 0.1031\textcolor{blue}{\ddag} & 0.2289 ± 0.1944\textcolor{blue}{\ddag} & 0.1324 ± 0.1391\textcolor{blue}{\ddag}\\
\midrule

\multirow{5}{*}{PEMS-SF} & $\checkmark$ & $\checkmark$ & $\checkmark$ & 0.5780 ± 0.0164\textcolor{white}{\ddag} & 0.6215 ± 0.0398\textcolor{white}{\ddag} & 0.5316 ± 0.0347\textcolor{white}{\ddag} & 0.3611 ± 0.0277\textcolor{white}{\ddag}\\
&  & $\checkmark$ & $\checkmark$ & 0.5318 ± 0.0216\textcolor{blue}{\ddag} & 0.5655 ± 0.0464\textcolor{blue}{\ddag} & 0.4771 ± 0.0146\textcolor{blue}{\ddag} & 0.3075 ± 0.0121\textcolor{blue}{\ddag}\\
& $\checkmark$ & & $\checkmark$ & 0.4229 ± 0.0661\textcolor{blue}{\ddag} & 0.6356 ± 0.0964\textcolor{white}{\ddag} & 0.5301 ± 0.0492\textcolor{blue}{\ddag} & 0.3735 ± 0.0581\textcolor{white}{\ddag} \\
&  & & $\checkmark$ & 0.5530 ± 0.0553\textcolor{blue}{\ddag} & 0.5632 ± 0.0499\textcolor{blue}{\ddag} & 0.4991 ± 0.0469\textcolor{blue}{\ddag} & 0.3441 ± 0.0571\textcolor{blue}{\ddag} \\
& $\checkmark$ & $\checkmark$ &  & 0.5954 ± 0.0805\textcolor{white}{\ddag} & 0.6237 ± 0.1035\textcolor{white}{\ddag} & 0.5690 ± 0.0829\textcolor{white}{\ddag} & 0.4229 ± 0.0661\textcolor{white}{\ddag}\\
\midrule

\multirow{5}{*}{RacketSports} & $\checkmark$ & $\checkmark$ & $\checkmark$ & 0.4715 ± 0.0135\textcolor{white}{\ddag} & 0.4657 ± 0.0695\textcolor{white}{\ddag} & 0.2004 ± 0.0520\textcolor{white}{\ddag} & 0.1235 ± 0.0372\textcolor{white}{\ddag}\\
&  & $\checkmark$ & $\checkmark$ & 0.4539 ± 0.0269\textcolor{blue}{\ddag} & 0.4477 ± 0.0558\textcolor{blue}{\ddag} & 0.1888 ± 0.0611\textcolor{blue}{\ddag} & 0.1284 ± 0.0623\textcolor{white}{\ddag}\\
& $\checkmark$ & & $\checkmark$ & 0.4737 ± 0.0352\textcolor{white}{\ddag} & 0.5160 ± 0.0655\textcolor{white}{\ddag} & 0.1881 ± 0.0461\textcolor{blue}{\ddag} & 0.1881 ± 0.0461\textcolor{white}{\ddag} \\
&  & & $\checkmark$ & 0.4539 ± 0.0215\textcolor{blue}{\ddag} & 0.4814 ± 0.0595\textcolor{white}{\ddag} & 0.1659 ± 0.0565\textcolor{blue}{\ddag} & 0.1177 ± 0.0322\textcolor{blue}{\ddag} \\
& $\checkmark$ & $\checkmark$ &  & 0.4496 ± 0.0248\textcolor{blue}{\ddag} & 0.4573 ± 0.0533\textcolor{blue}{\ddag} & 0.1441 ± 0.0530\textcolor{blue}{\ddag} & 0.1005 ± 0.0512\textcolor{blue}{\ddag}\\
\midrule

\multirow{5}{*}{\shortstack[l]{SelfRegulation\\SCP1}} & $\checkmark$ & $\checkmark$ & $\checkmark$ & 0.8510 ± 0.0126\textcolor{white}{\ddag} & 0.8500 ± 0.0135\textcolor{white}{\ddag} & 0.4146 ± 0.0207\textcolor{white}{\ddag} & 0.4916 ± 0.0355\textcolor{white}{\ddag}\\
&  & $\checkmark$ & $\checkmark$ & 0.8248 ± 0.0253\textcolor{blue}{\ddag} & 0.8235 ± 0.0250\textcolor{blue}{\ddag} & 0.3591 ± 0.0680\textcolor{blue}{\ddag} & 0.4226 ± 0.0653\textcolor{blue}{\ddag}\\
& $\checkmark$ & & $\checkmark$ & 0.8111 ± 0.0485\textcolor{blue}{\ddag} & 0.8075 ± 0.0523\textcolor{blue}{\ddag} & 0.3423 ± 0.0774\textcolor{blue}{\ddag} & 0.3947 ± 0.1155\textcolor{blue}{\ddag} \\
&  & & $\checkmark$ & 0.8134 ± 0.0265\textcolor{blue}{\ddag} & 0.8133 ± 0.0265\textcolor{blue}{\ddag} & 0.3112 ± 0.0567\textcolor{blue}{\ddag} & 0.3937 ± 0.0668\textcolor{blue}{\ddag} \\
& $\checkmark$ & $\checkmark$ &  & 0.7770 ± 0.0607\textcolor{blue}{\ddag} & 0.7756 ± 0.0611\textcolor{blue}{\ddag} & 0.2628 ± 0.1234\textcolor{blue}{\ddag} & 0.3194 ± 0.1427\textcolor{blue}{\ddag}\\
\midrule

\multirow{5}{*}{\shortstack[l]{SelfRegulation\\SCP2}} & $\checkmark$ & $\checkmark$ & $\checkmark$ & 0.6000 ± 0.0253\textcolor{white}{\ddag} & 0.6000 ± 0.0253\textcolor{white}{\ddag} & 0.0403 ± 0.0221\textcolor{white}{\ddag} & 0.0380 ± 0.0197\textcolor{white}{\ddag}\\
&  & $\checkmark$ & $\checkmark$ & 0.5889 ± 0.0120\textcolor{blue}{\ddag} & 0.5819 ± 0.0143\textcolor{blue}{\ddag} & 0.0257 ± 0.0070\textcolor{blue}{\ddag} & 0.0271 ± 0.0089\textcolor{blue}{\ddag}\\
& $\checkmark$ & & $\checkmark$ & 0.6000 ± 0.0136\textcolor{white}{\ddag} & 0.5987 ± 0.0134\textcolor{blue}{\ddag} & 0.0302 ± 0.0083\textcolor{blue}{\ddag} & 0.0354 ± 0.0110\textcolor{blue}{\ddag} \\
&  & & $\checkmark$ & 0.6019 ± 0.0105\textcolor{white}{\ddag} & 0.5915 ± 0.0181\textcolor{blue}{\ddag} & 0.0350 ± 0.0046\textcolor{blue}{\ddag} & 0.0371 ± 0.0086\textcolor{blue}{\ddag} \\
& $\checkmark$ & $\checkmark$ &  & 0.5667 ± 0.0157\textcolor{blue}{\ddag} & 0.5453 ± 0.0210\textcolor{blue}{\ddag} & 0.0195 ± 0.0105\textcolor{blue}{\ddag} & 0.0143 ± 0.0078\textcolor{blue}{\ddag}\\
\midrule

\multirow{5}{*}{StandWalkJump} & $\checkmark$ & $\checkmark$ & $\checkmark$ & 0.7556 ± 0.0831\textcolor{white}{\ddag} & 0.7382 ± 0.1048\textcolor{white}{\ddag} & 0.5189 ± 0.1349\textcolor{white}{\ddag} & 0.3709 ± 0.1810\textcolor{white}{\ddag}\\
&  & $\checkmark$ & $\checkmark$ & 0.6667 ± 0.0544\textcolor{blue}{\ddag} & 0.6536 ± 0.0599\textcolor{blue}{\ddag} & 0.4238 ± 0.0697\textcolor{blue}{\ddag} & 0.2191 ± 0.1006\textcolor{blue}{\ddag}\\
& $\checkmark$ & & $\checkmark$ & 0.6667 ± 0.0544\textcolor{blue}{\ddag} & 0.6553 ± 0.0755\textcolor{blue}{\ddag} & 0.3612 ± 0.0928\textcolor{blue}{\ddag} & 0.1637 ± 0.0649\textcolor{blue}{\ddag} \\
&  & & $\checkmark$ & 0.6889 ± 0.0314\textcolor{blue}{\ddag} & 0.6677 ± 0.0531\textcolor{blue}{\ddag} & 0.3845 ± 0.0815\textcolor{blue}{\ddag} & 0.2118 ± 0.0444\textcolor{blue}{\ddag} \\
& $\checkmark$ & $\checkmark$ &  & 0.6222 ± 0.0629\textcolor{blue}{\ddag} & 0.5976 ± 0.0636\textcolor{blue}{\ddag} & 0.4059 ± 0.1210\textcolor{blue}{\ddag} & 0.2214 ± 0.1416\textcolor{blue}{\ddag}\\
\bottomrule
\end{tabular}
}
\caption{Ablation study of loss terms. The symbol ``$\checkmark$'' denotes included terms, and the symbol ``\textcolor{blue}{\ddag}'' indicates performance degradation compared to the complete EMTC.}
\label{tb:ablation_loss}
\end{table}

\subsection{Ablation Study of Loss Terms}
The results in Table~\ref{tb:ablation_loss} demonstrate that all three loss terms are essential for EMTC's optimal performance. The complete model with the full loss combination consistently outperforms all ablated variants, with the absence of any loss component causing noticeable performance degradation. The contrastive loss generally contributes most substantially to clustering quality, particularly on complex multi-class datasets like Cricket where cluster separation is challenging. It is worth noting that the performance degradation becomes more pronounced when multiple loss terms are removed simultaneously, confirming the complementary nature of the reconstruction and contrastive learning objectives. Overall, the ablation validates the synergistic design of the loss function: $\mathcal{L}_{\text{contra}}$ enhances cluster discrimination in the embedding space, while $\mathcal{L}_{\text{intra}}$ and $\mathcal{L}_{\text{inter}}$ together ensure robust MEV representation learning against masked redundancy.

\subsection{Efficiency Analysis}
\begin{figure}[!t]
\centering
\includegraphics[width=\columnwidth]{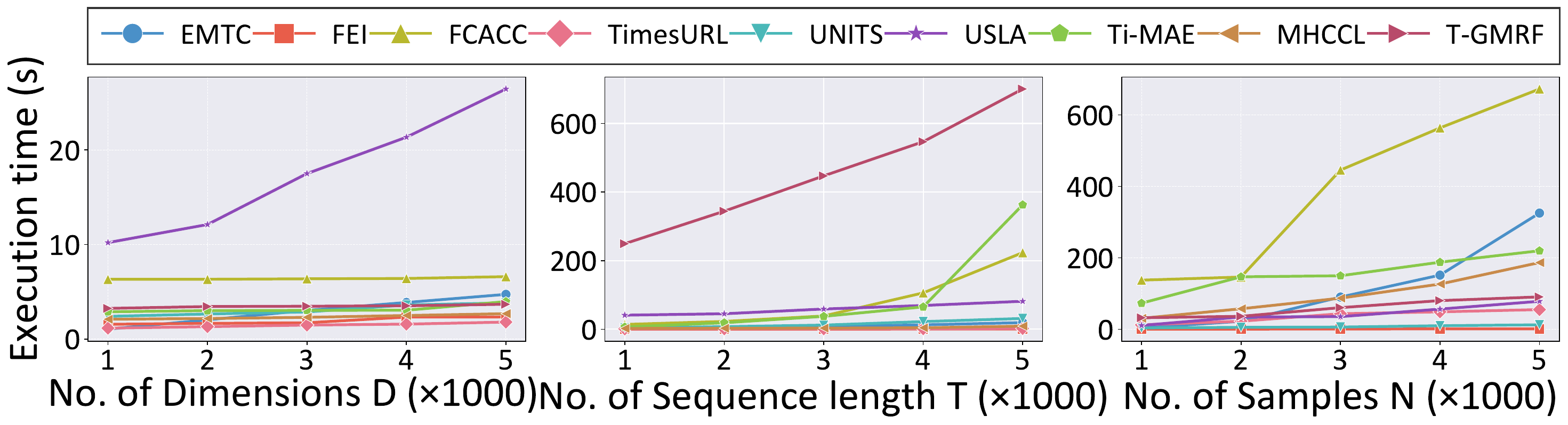}
\caption{Execution time (s) of model training on synthetic datasets with varying $D$, $T$, and $N$.}
\label{time}
\end{figure}

Efficiency and scalability of EMTC are evaluated by recording its execution time with respect to three key factors: number of dimensions ($D$), sequence length ($T$), and number of samples ($N$). As shown in Figure~\ref{time}, EMTC demonstrates favorable scalability across different dimensions. When $D$ and $T$ scale, EMTC exhibits near-linear growth, maintaining low computational overhead.

\subsection{Cluster Effect Visualization}
\begin{figure}[!t]
\centering
\includegraphics[width=\columnwidth]{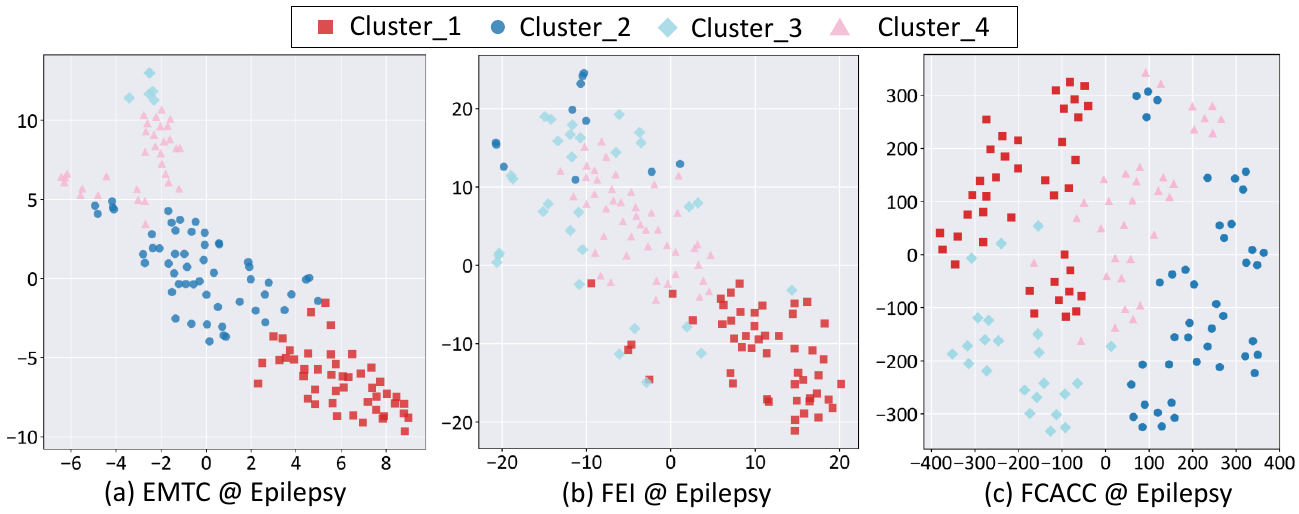}
\caption{t-SNE visualization of the Epilepsy dataset represented by different methods.}
\label{visual}
\end{figure}

To qualitatively assess representation quality from the perspective of clustering, Figure~\ref{visual} presents t-SNE visualizations of learned representations on the Epilepsy dataset, comparing EMTC with FEI and FCACC. EMTC demonstrates superior cluster separation with relatively prominent boundaries and compact groupings, while FEI exhibits substantial inter-cluster overlap and FCACC shows moderate overlap but with its clusters scattered. These visual results confirm that the EMTC effectively captures clustering-friendly patterns from MTS, yielding semantically meaningful embeddings for MTS clustering.

\section{Concluding Remarks}
This paper proposes EMTC, an evolving-masking guided MTS clustering framework that explicitly addresses temporal redundancy through learnable masking and multi-view representation learning. It introduces IVM module to dynamically suppress redundant timestamps via attention-based scoring, ensures that the representation learning focuses on discriminative temporal regions. A dual-path MEV learning is also adopted, integrating the CRL and CMC learning to ensure the effectiveness of IVM and facilitate clustering-friendly representation learning, respectively. EMTC is comprehensively validated on 15 benchmark datasets with comparative, ablation, efficiency, and qualitative studies, demonstrating its superior performance over the SOTA methods. To our knowledge, EMTC is the first dynamic masking attempt under the scenario of unsupervised MTS analysis, breaking the clustering performance bottleneck caused by the MTS redundancy, and providing a mask learning paradigm to unsupervised MTS analysis. Despite the above-mentioned merits, this work is also not exempt from limitations. IVM relies on determining a to-be-masked threshold, while MEV generation assumes that all data is known. Extending EMTC to automatically determine the masking threshold and continuously adapt to nonstationary streaming data would be the next avenue.

\section{Acknowledgment}
This work was supported in part by the National Natural Science Foundation of China under Grant 62476063 and the Natural Science Foundation of Guangdong Province under Grant 2025A1515011293.

\bibliography{aaai2026}
\end{document}